\documentclass[runningheads]{llncs}

\usepackage{booktabs}
\usepackage[table,xcdraw]{xcolor}
\usepackage{graphicx}
\usepackage{subcaption}
\usepackage{cite}
\usepackage{svg}
\usepackage[hidelinks]{hyperref}

\usepackage{comment}
\usepackage{caption}
\usepackage{multirow}
\usepackage{graphicx}

\makeatletter
\newcommand\footnoteref[1]{\protected@xdef\@thefnmark{\ref{#1}}\@footnotemark}
\makeatother

\begin{document}
\title{Egocentric Human-Object Interaction Detection\\ Exploiting Synthetic Data}
\author{Rosario Leonardi\inst{1} \and Francesco Ragusa\inst{1, 2} \and \\ Antonino Furnari\inst{1, 2} \and Giovanni Maria Farinella\inst{1, 2}}
\authorrunning{R. Leonardi et al.}
\institute{FPV@IPLAB, DMI - University of Catania, Italy \and  Next Vision s.r.l. - Spinoff of the University of Catania, Italy}
\maketitle
\begin{abstract}
We consider the problem of detecting Egocentric Human-Object Interactions (EHOIs) in industrial contexts. Since collecting and labeling large amounts of real images is challenging, we propose a pipeline and a tool to generate photo-realistic synthetic First Person Vision (FPV) images automatically labeled for EHOI detection in a specific industrial scenario. To tackle the problem of EHOI detection, we propose a method that detects the hands, the objects in the scene, and determines which objects are currently involved in an interaction. We compare the performance of our method with a set of state-of-the-art baselines. Results show that using a synthetic dataset improves the performance of an EHOI detection system, especially when few real data are available. To encourage research on this topic, we publicly release the proposed dataset at the following url: \url{https://iplab.dmi.unict.it/EHOI_SYNTH/}.
\end{abstract}

\section{Introduction}
Understanding Human-Object Interactions (HOI) from the first-person perspective allows to build intelligent systems able to understand how humans interact with the world. The use of wearable cameras can be highly relevant to understand users' locations of interest~\cite{Furnari2016474}, to assist visitors in cultural sites~\cite{CucchiaraVACHE, Fragusa_EVLCS}, to provide assistance to people with disabilities~\cite{wang2019learning}, or to improve the safety of workers in a factory~\cite{Ragusa2021TheMD}. %In particular, in the industrial domain, an intelligent FPV assistant can be used to monitor human-object interactions, verify the correctness of an interaction in a know workflow, and train users on how to interact with some specific objects.
Despite the rapid growth of wearable devices~\cite{Betancourt_evolution_fpv_methods}, the task of Egocentric Human-Object Interaction (EHOI) detection is still understudied in this domain due to the limited availability of public datasets~\cite{Ragusa2021TheMD}. %Indeed, data acquisition in industrial environments is difficult due to privacy concerns and the need to preserve industrial secrets. \\ 
%In this paper, we consider the problem of detecting and recognizing EHOIs in an industrial context. 
%In recent years, many works focused on HOI detection by considering the third person view. The work of~\cite{GkioxariDRHOI} defines HOI as the task of detecting a \textit{\textless human, verb, object\textgreater} triple from an image.
%Since in FPV only the hands of the interacting agent are generally visible, in this setting the definition of~\cite{Shan2020UnderstandingHH} can be considered. More specifically in FPV, an interaction can be described as \textit{\textless hand, contact\_state, object\textgreater} triple, where the ``\textit{contact\_state}'' variable assumes one of the following values: \textit{(none/self/other/portable/non-portable)}. 
%Given the relevance of wearable devices to industrial contexts, we consider First Person Vision (FPV) settings as in~\cite{Ragusa2021TheMD}. 
We note that in an industrial domain, in which the set of objects of interest is known a priori (e.g., the tools and instruments the user is going to interact with), the ability to detect the user's hands, find all objects and determine which objects are involved in an interaction, can inform on the user's behavior and provide useful information for other tasks such as object interaction anticipation~\cite{AFurnari_next_active, furnari2020rulstm}.
%We consider First Person Vision (FPV) settings as in~\cite{Ragusa2021TheMD} and focus on two classes of contact state~\cite{Shan2020UnderstandingHH}: \textit{In contact} vs \textit{No contact}. We note that in an industrial domain, in which the set of objects of interest is known a priori (e.g., the tools and instruments the user is going to interact with), the ability to detect the objects the user interacts with \textit{(in contact)}, can provide useful information for other tasks such as object interaction anticipation~\cite{AFurnari_next_active, furnari2020rulstm}.
Extending the definition proposed in~\cite{Shan2020UnderstandingHH}, we hence consider the problem of detecting an EHOI as the one of predicting a quadruple \textit{\textless hand, contact\_state, active\_object, \textless other\_objects\textgreater\textgreater}.
%We extend the definition of~\cite{Shan2020UnderstandingHH} to also include the detection of all the objects in the scene, thus obtaining a quadruple \textit{\textless hand, contact\_state, active\_object, \textless other\_objects\textgreater\textgreater}. 

\begin{figure}[t]
    \centering
    \includegraphics[width=\textwidth]{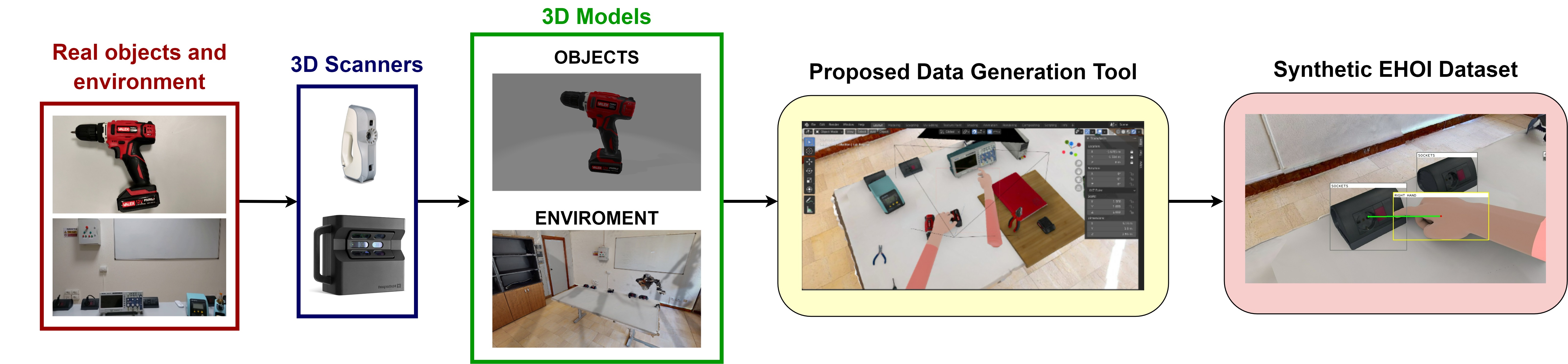}
    \caption{Synthetic EHOIs generation pipeline. We first use 3D scanners to obtain 3D models of the set of objects and the environment. We hence use the proposed data generation tool to create the synthetic EHOI dataset.}
    \label{fig_data_generation_pipeline}
\end{figure}

\begin{figure}[t]
    \centering
    \includegraphics[width=0.8\textwidth]{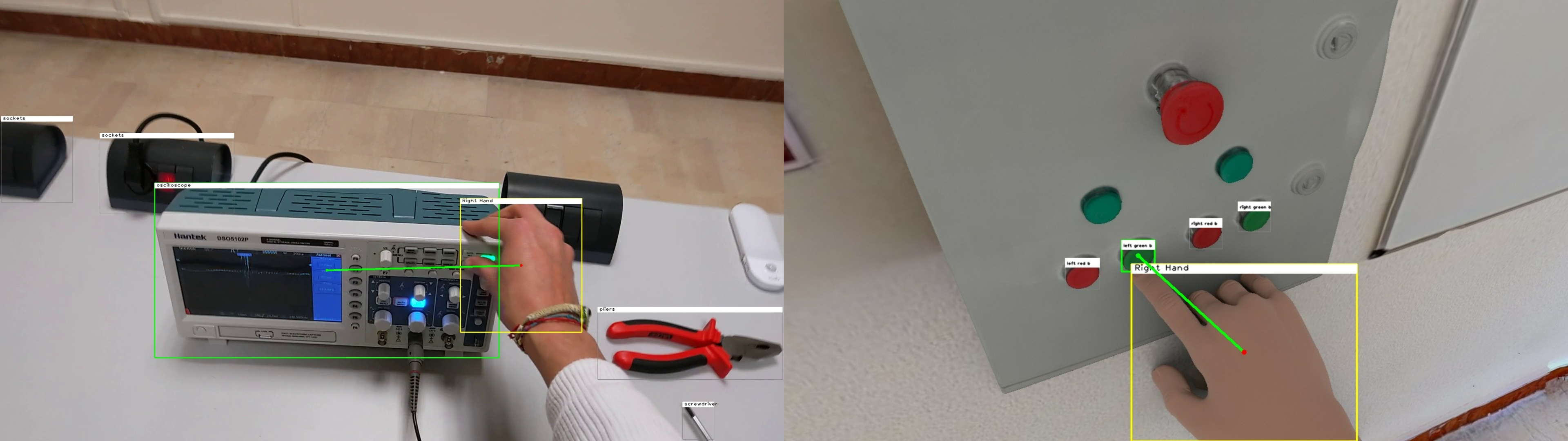}
    \caption{Example output of the proposed system. The figure on the left shows an example of a real image whereas a synthetic image is shown on the right. %The method recognizes all the objects in the scene, the objects that are involved in an interaction (indicated with green bounding boxes), and detects the hands. Furthermore, the method recognizes the side of each hand (\textit{Right} or \textit{Left}) and the contact state (\textit{In contact} or \textit{No contact}). Lastly, the system associates each hand with the object the user is interacting with (indicated by a green line in the figure).
    } 
    \label{fig_desired_output}
\end{figure}

To develop a system able to tackle this task in a specific industrial scenario, it is generally required to collect and label large amounts of data. %Indeed, for each frame, it is required to label all the hands, including attributes (e.g. side, or the contact state), the active and non-active objects, including the objects categories and bounding boxes, and the relationship between the hands and the active objects.
To reduce the significant costs usually required for data collection and annotation, we investigated whether the use of synthetic images 
%Since the data collection and annotation phases usually require a substantial amount of time and cost, we investigated whether the use of synthetic images %, automatically generated and labeled from 3D models of the environment and objects,
can help to achieve good performance when models are trained on synthetic data and tested on real one. To this end, we propose a pipeline and a tool to generate a large number of synthetic EHOIs from 3D models of a real environment and objects. Unlike previous approaches~\cite{HassonLJR, Mueller2018GANeratedHF}, we generate EHOIs simulating a photo-realistic industrial environment. The proposed pipeline (Figure \ref{fig_data_generation_pipeline}) allows to obtain 3D models of the objects and the environment using 3D scanners. Such models are then used with the proposed data generation tool to automatically produce labeled images of EHOIs. Even though some works provide datasets to study HOI in general domains~\cite{Shan2020UnderstandingHH, Damen2018ScalingEV, Damen2021RESCALING, Bambach2015LendingAH, Lu2021TheOA} and in industrial contexts~\cite{Ragusa2021TheMD} %, grauman2021ego4d}, 
 to the best of our knowledge, this is the first attempt to define a pipeline for the generation of a large-scale photo-realistic synthetic FPV dataset to study EHOIs with rich annotations of active and non-active objects in an industrial scenario.
To assess the suitability of the generated synthetic data to tackle the EHOI detection task, we acquired and labeled 8 real egocentric videos in an industrial laboratory, in which subjects perform test and repair operations on electrical boards (see Figure~\ref{fig_desired_output}). %We annotated all instances of EHOIs in these real data indicating the frames in which they occur and all the involved active objects (i.e., the objects to the user is interacting with) with a bounding box associated with a class label. In addition, we labeled the hands and all the other non-active objects. \\
%\indent To test the usefulness of the generated synthetic data to tackle the EHOI detection task, we acquired 8 real videos, of about 25 minutes long on average, using a Microsoft Hololens 2\footnote{\hyperlink{https://www.microsoft.com/en-us/hololens}{https://www.microsoft.com/en-us/hololens}} device. Data have been acquired in an laboratory reproducing an industrial enviroment in which subjects perform test and repair operations on electrical boards (See Figure~\ref{fig_desired_output}). We annotated all instances of EHOIs in these real data indicating the frames in which they occur and all the involved active objects (i.e., the objects to the user is interacting with) with a bounding box associated with a class label. In addition, we labeled the hands and all the other non-active objects. \\
To address the problem of EHOI detection, we propose a method inspired by~\cite{Shan2020UnderstandingHH} that detects and recognizes all the objects in the scene, determining which of these are involved in an interaction, as well as the hands of the camera wearer (see Figure \ref{fig_desired_output}). %The method determines which objects are currently involved in an interaction.
%Differently than~\cite{Shan2020UnderstandingHH} the proposed approach aims to infer also the class of the active objects.
%An example of the desired output from a model is shown in Figure \ref{fig_desired_output}.
To investigate the usefulness of exploiting synthetic data for EHOIs detection, we trained the proposed method using all the synthetic images together with variable amounts of real data. %(0\%, 10\%, 25\%, 50\%, 100\%).
In addition, we compared the results of the proposed approach with different instances of the method proposed in~\cite{Shan2020UnderstandingHH}. The results show that using synthetic data improves the performance of the EHOI method when tested on real images.

In sum, the contributions of this paper are as follows: 1) we present a new photo-realistic synthetic FPV dataset for EHOIs detection considering an industrial scenario with rich annotations of the hands, and the active/non-active objects, including class labels and semantic segmentation masks; 2) we propose a method inspired by~\cite{Shan2020UnderstandingHH} which detects and recognizes all the objects in the scene, the hands of the camera wearer, and determines which objects are currently involved in an interaction;
3) we perform several experiments to investigate the usefulness of synthetic data for the EHOI detection task when the method is tested on real data and compare the obtained results with baseline approaches based on the state-of-the-art method described in~\cite{Shan2020UnderstandingHH}.
%\noindent The remainder of the paper is organized as follows. In Section \ref{sec_related_work} we discuss related work. The proposed dataset is presented in Section \ref{sec_dataset}. Section \ref{sec_method} describes the proposed approach. Section \ref{sec_experiments_results} reports the experimental settings and discusses the results. We conclude the paper and summarize the main results of our study in Section \ref{sec_conclusion}.

\section{Related Work}\label{sec_related_work}
\paragraph{Datasets for Human Behavior Understanding} In recent years, many works focused on the Human-Object Interaction detection task considering the third-person point of view. Several datasets have been proposed to explore this problem. The authors of~\cite{Gupta2015VisualSR} proposed the V-COCO dataset, which adds 26 verbs to the 80 object classes of the popular COCO dataset~\cite{coco_dataset}. HICO-DET~\cite{chao:wacv2018} includes over 600 distinct interaction classes, while HOI-A~\cite{Liao2020PPDMPP} considers 10 action categories and 11 object classes. Previous works have also proposed datasets of videos to address the action recognition task. We can mention the ActivityNet dataset~\cite{Heilbron2015ActivityNetAL} which focuses on 200 different action classes as well as Kinetics~\cite{kay2017kinetics} which contains over 700 human action classes. With the rapid growth of wearable devices, different datasets of images and videos captured from the first-person point of view have been proposed. Among these, the work of~\cite{Bambach2015LendingAH} provided a dataset of 48 FPV videos of people interacting with objects, including segmentation masks for 15,000 hand instances. EPIC-Kitchens~\cite{Damen2021RESCALING, Damen2018ScalingEV} is a series of egocentric datasets focused on unscripted activities of human behavior in kitchens. EGTEA Gaze+~\cite{li2020eye} is a dataset of 28 hours of video of cooking activities. The dataset 100 Days of Hands (100DOH)~\cite{Shan2020UnderstandingHH} is composed of both Third Person Vision (TPV) and FPV images and is suitable to study object-class agnostic HOI detection. The authors of~\cite{Lu2021TheOA} labeled images collected from different FPV datasets~\cite{Damen2018ScalingEV, Li2015DelvingIE, GarciaHernando2018FirstPersonHA} providing annotations for hands, objects and their relation. The MECCANO dataset~\cite{Ragusa2021TheMD} contains videos acquired in an industrial-like domain also annotated with bounding boxes around active objects, together with the related classes. A massive-scale egocentric dataset named Ego4D\footnote{Ego4D Website: \url{https://ego4d-data.org/}} has been acquired in various domains and labeled with several annotations to address different challenges. Since the annotation phase of EHOIs is expensive in terms of costs and time, the use of synthetic datasets for training purposes is desired. A few works explored the use of synthetic images generated from the first-person point of view~\cite{HassonLJR, Mueller_real_time_hand_tracking_2017}. The authors of such works used different strategies to customize various aspects of the scene, such as lights and backgrounds. However, these approaches tend to produce non-photorealistic images.
Differently from the aforementioned works, we generate a dataset of photo-realistic synthetic images of EHOIs in an industrial environment with rich annotations of hands, including hand side (\textit{Left/Right}), contact state (\textit{In contact with an object/No contact}), and all the objects in the images with bounding boxes. We also provide a class label for each object and indicate whether it is an active object as well as semantic segmentation masks.

\paragraph{Understanding Human-Object Interactions} 
There has been a lot of research in computer vision focusing on understanding Human-Object Interactions. The work of~\cite{GkioxariDRHOI} presented a multitask learning system to tackle HOI detection. The proposed system consists of an object detection branch, a human-centric branch, and an interaction branch. The authors of~\cite{Shan2020UnderstandingHH} tackled the HOI detection task from both TPV and FPV predicting different information about hands (i.e., bounding box, hand side and contact state) and a box around the interacted object. PPDM~\cite{Liao2020PPDMPP} defines an HOI as a point triplet \textit{\textless human point, interaction point, object point\textgreater} where these points represent the center of the related bounding boxes. The authors of~\cite{Zhang2021EfficientTD} proposed a new two-stage detector called Unary–Pairwise Transformer. This approach exploits unary and pairwise representations to detect Human-Object Interactions. However, all these works mainly consider third-person view scenarios. Indeed, this task is still understudied in the FPV domain. Previous FPV works focused on the detection of hands interacting with an object without recognizing it~\cite{Bambach2015LendingAH, Lu2021UnderstandingEH}. Other recent works focused on object-class agnostic EHOI detection~\cite{Lu2021TheOA, Fu2021SequentialDF}. The authors of~\cite{Ragusa2021TheMD} defined Egocentric Human-Object Interaction (EHOI) detection as the task of producing \textit{\textless verb, objects\textgreater} pairs. The paper investigated the problem of recognizing active objects in industrial-like settings without considering hands. The authors of~\cite{chen2019holistic} considered the usage of synthetic data for recognizing the performed Human-Objects Interactions.
In this paper, we tackle the EHOIs detection task in an industrial domain and investigate the usefulness of using synthetic data for training when the system needs to be tested on real data. In addition, our approach aims to detect both active and non-active objects as well as infer their classes.

\section{Dataset}\label{sec_dataset}
\begin{figure}[t]
    \centering
    \includegraphics[width=0.8\textwidth]{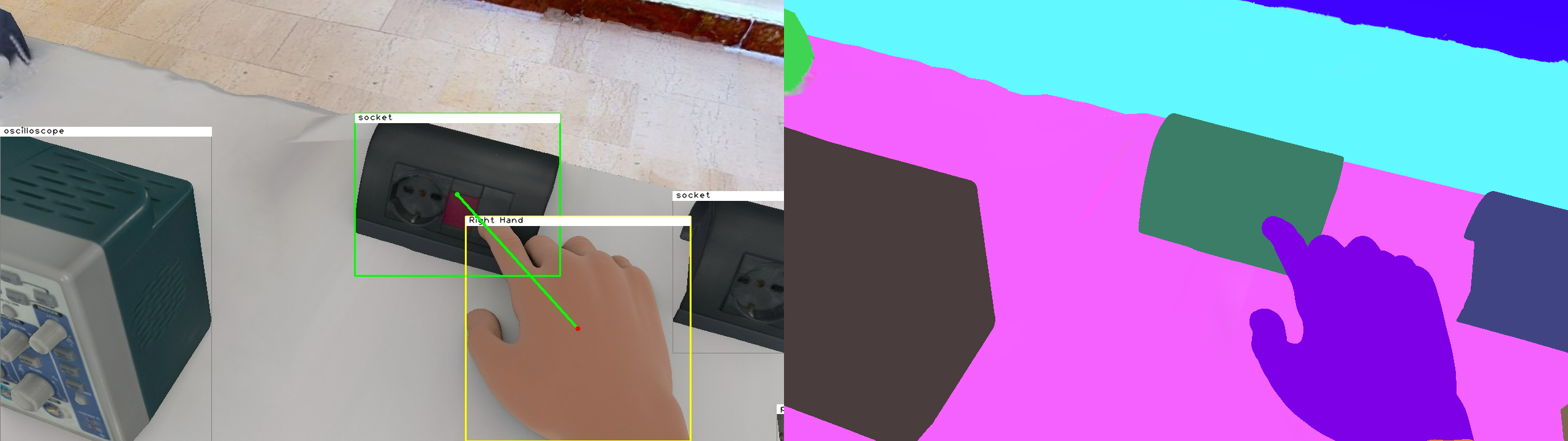}
    \caption{Example of a synthetic EHOI generated with the developed tool. On the left, the figure shows the synthetic RGB image automatically labeled (left) %along with annotations related to the hands, the objects, and the relation between hands and active objects.
    as well as the semantic segmentation mask generated for the same EHOI (right).}
    \label{fig_synthetic_image_example}
\end{figure}

\paragraph{Industrial context} 
We set up a laboratory to study the EHOIs detection task in a realistic industrial context. In the considered laboratory there are different objects, such as a power supply, a welding station, sockets, and a screwdriver. In addition, there is an electrical panel that allows powering on and off the sockets\footnote{\label{seesup}See supplementary material for more details.}.
To generate synthetic data compliant to the considered real space, we acquire 3D scans of all objects and of the environment. It is worth noting that for the small objects, high-quality reconstructions are required to generate realistic EHOIs, whereas for the reconstruction of the environment, a high accuracy is not needed. Hence, to create 3D models, we used two different 3D scanners. In particular, we used an Artec Eva\footnote{\url{https://www.artec3d.com/portable-3d-scanners/artec-eva-v2}} structured-light 3D scanner, which has a 3D resolution of up to 0.2 mm, to scan the objects, and a MatterPort\footnote{\url{https://matterport.com/}} device to scan the 3D model of the environment.
%Figure \ref{fig_industrial_laboratory} shows the laboratory. \\
%\begin{figure}[t]
%    \centering
%    \includegraphics[width=\textwidth]{images/industrial_lab_combo.jpg}
%    \caption{Industrial-like laboratory.}
%    \label{fig_industrial_laboratory}
%\end{figure}
\paragraph{Synthetic Data}
We adopted the pipeline shown in Figure \ref{fig_data_generation_pipeline} to generate the synthetic data of EHOIs in the considered industrial context. We developed a tool in Blender which takes as input the 3D models of the objects and the environment and generates synthetic EHOIs along with different data, including 1) photo-realistic RGB images (see Figure \ref{fig_synthetic_image_example} - left), 2) depth maps, 3) semantic segmentation masks (see Figure \ref{fig_synthetic_image_example} - right), 4) objects bounding boxes and categories indicating which of them are active, 5) hands bounding boxes and attributes, such as the hand side (\textit{Left/Right}) and the contact state (\textit{In contact with an object/No contact}), and 6) distance between hands and objects in the 3D space. The tool allows to customize different aspects of the virtual scene, including the camera position, the lighting, and the color of the hands for automatic acquisition. Figure \ref{fig_synthetic_image_example} shows an example of synthetic EHOIs and related labels generated with our tool.
The generated synthetic dataset contains a total of 20,000 images, 29,034 hands (of which 14,589 involved in an interaction), 123,827 object instances (14,589 of which are active objects), and 19 object categories including portable industrial tools (e.g., screwdriver, electrical boards) and instruments (e.g., power supply, oscilloscope, electrical panels)\footnoteref{seesup}.%\footnote{See the supplementary material for additional details.}. %Additional details are provided in the supplementary material. % Figure \ref{fig_statistics_number_instances} shows the distribution of the active objects. Table \ref{tab_statistics_dataset_synthetic} reports the statistics related to the generated dataset. 
\begin{comment}
\begin{table}[t]
	\begin{minipage}{0.6\linewidth}
		\centering
        \includegraphics[width=1\textwidth]{images/Number of class instances active.pdf}
		\captionof{figure}{Distribution of active objects instances in the synthetic dataset.}
        \label{fig_statistics_number_instances}
	\end{minipage}\hfill
	\begin{minipage}[b]{0.38\linewidth}
		\centering
        \caption{Statistics of the generated synthetic dataset.}
        \label{tab_statistics_dataset_synthetic}
        \resizebox{\linewidth}{!}{%
        \begin{tabular}{l|c}
            \textbf{Total number of images} & 20,000    \\ \hline
            \textbf{\#hands}                & 29,034    \\
            \textbf{\#hands in contact}     & 14,589    \\
            \textbf{\#hands not in contact} & 14,445    \\
            \textbf{\#left hands}           & 14,473    \\
            \textbf{\#right hands}          & 14,561    \\
            \textbf{\#object categories}    & 19        \\ 
            \textbf{\#objects}              & 123,827   \\
            \textbf{\#active objects}       & 14,589    \\ \hline
        \end{tabular} 
        }
	\end{minipage}
\end{table}
\end{comment}
\begin{figure}[t]
    \centering
    \includegraphics[width=0.85\textwidth]{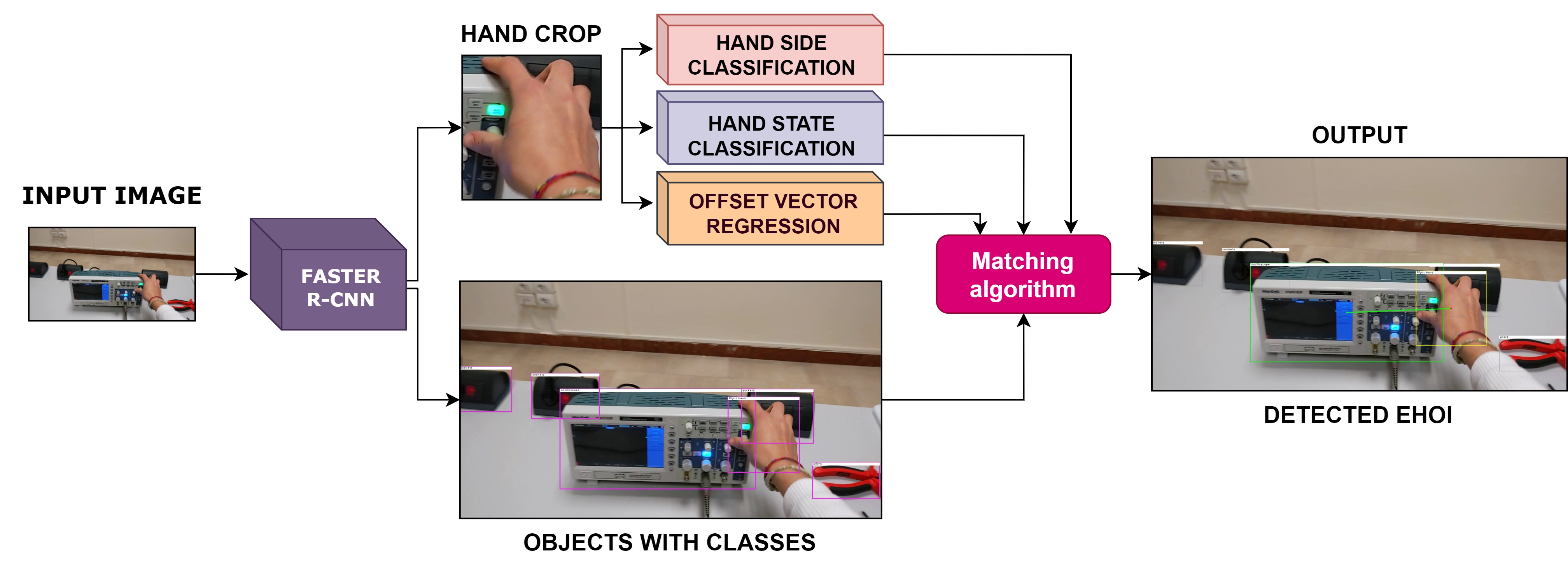}
    \caption{The proposed system takes an egocentric RGB image as input and outputs several predictions about the status of hands and objects involved in the interactions.} \label{fig_schema_rete}
\end{figure}
\paragraph{Real Data}
The real data consists in 8 real videos acquired using a Microsoft Hololens 2 wearable device. To this aim, we asked 7 different subjects to perform test and repair operations on electrical boards in the industrial laboratory. To simplify the acquisition process, we defined different sequences of operations that subjects have to follow (e.g., turning on the oscilloscope, connecting the power cables to the electrical board, etc). To make data collection consistent and more natural, we developed a Mixed-Reality application for Hololens 2 that guides the subjects through audio and images to the next operation they have to perform. The set of operations has been randomized in order to be less scripted. The average duration of the captured videos is 28.37 minutes. In total, we acquired 3 hours and 47 minutes of video recordings at a resolution of 2272x1278 pixels and with a framerate of 30fps. An example of the captured data is shown in Figure \ref{fig_desired_output} - left. We manually annotated the real videos with all the EHOIs performed by the subjects. We used the following approach to select the image frames to be annotated: 1) we considered the first frame in which the hand touches the interacted object (i.e., contact frame), and 2) we selected the first frame that appears immediately after the hand released the object (i.e., non contact frame). For each of the considered frames we annotated: 1) hand bounding boxes and attributes, such as hand side and contact state \textit{(In Contact with an object/No contact)}, 2) active and non-active object bounding boxes and their categories, and 3) the relationships between the hands and the active objects (e.g. \textit{in contact with the right hand})\footnoteref{seesup}. %\footnote{See the supplementary material for additional details.}. %Additional details are provided in the supplementary material. %The distribution of the active object instances is shown in Figure \ref{fig_statistics_number_instances_real} whereas Table \ref{tab_statistics_real_dataset} reports the statistics of the real dataset.
\begin{comment}
\begin{table}[t]
	\begin{minipage}{0.6\linewidth}
		\centering
        \includegraphics[width=1\textwidth]{images/Number_of_class_instances_active real.pdf}
		\captionof{figure}{Distribution of the active objects instances in the real data.}
        \label{fig_statistics_number_instances_real}
	\end{minipage}\hfill
	\begin{minipage}[b]{0.38\linewidth}
        \centering
        \caption{Statistics of the real dataset.}
        \label{tab_statistics_real_dataset}
        \resizebox{\linewidth}{!}{%
        \begin{tabular}{l|c}
            \textbf{\#videos} & 8  \\ \hline
            \textbf{Total videos length} & 227 min \\
            \textbf{Avg. video duration} & 28.37 min \\
            \textbf{\#subjects} & 7  \\
            \textbf{\#images} & 3,056  \\
            \textbf{\#hands} & 4,503     \\
            \textbf{\#hands in contact} & 3,311 \\
            \textbf{\#hands not in contact} & 1,192  \\
            \textbf{\#left hands} & 2,013     \\
            \textbf{\#right hands} & 2,490  \\
            \textbf{\#object categories}  & 19 \\ 
            \textbf{\#objects} & 17,598 \\
            \textbf{\#active objects} & 2,872\\\hline
        \end{tabular} 
        }
	\end{minipage}
\end{table}
\end{comment}

\section{Proposed Approach}\label{sec_method}
Similarly to~\cite{Shan2020UnderstandingHH}, our method extends the popular two-stage detector Faster R-CNN~\cite{faster_rccn} to address the considered EHOIs detection task. However, differently than~\cite{Shan2020UnderstandingHH}, the proposed method is able to detect all the objects in the image together with the active/no active object class. Figure~\ref{fig_schema_rete} shows the architecture of the proposed approach. The proposed method detects the hands and the objects in an egocentric RGB image and infers: 1) object categories, 2) hands side, 3) hands contact state, and 4) EHOIs as \textit{\textless hand, contact\_state, active\_object, \textless other\_objects\textgreater\textgreater} quadruplet. Similarly to~\cite{Shan2020UnderstandingHH}, we extend the object detector with four additional components: 1) the hand side classification module, 2) the hand state classification module, 3) the offset vector regression module, and 4) the matching algorithm. The modules composing our method are described in the following\footnoteref{seesup}.
\\
%\footnote{See the supplementary material for training details.}. \\ 
\noindent\textbf{Hands and objects detection:} For objects and hands detection, we adopted a Faster R-CNN detector~\cite{faster_rccn} based on a ResNet-101 backbone~\cite{he2015deep} and a Feature Pyramid Network (FPN)~\cite{lin2017feature} due to their state-of-the-art performance. The network predicts a \textit{(x,y,w,h,c)} tuple for each object/hand in the image, where the \textit{(x,y,w,h)} tuple represents the bounding box coordinates, and \textit{c} is the predicted object class. \\
\noindent\textbf{Hand side classification module:} The hand side classification module consists of a Multi-Layer Perceptron (MLP) composed of two fully connected layers. Starting from the detected hands, it takes as input a ROI-pooled feature vector of the hand crop and predicts the side of the hand \textit{(left/right)}.\\
\noindent\textbf{Hand state classification module:} We consider two contact state classes: \textit{In~Contact} and \textit{No contact}. Other information about the contact state is embedded in the object category, which is predicted by our method, as opposed to~\cite{Shan2020UnderstandingHH} which predicts several types of contact states such as ``in contact with a mobile object'' or ``in contact with a fixed object''. The hand state classification module is composed of a MLP with two fully connected layers. We also enlarge the hand crop by 30\% relative to the detected bounding box to include information of the surrounding context (e.g., nearby objects). The module takes as input the ROI-pooled feature vectors to infer the hands contact state. \\ 
\noindent\textbf{Offset vector regression module:} Following the approach proposed in~\cite{Shan2020UnderstandingHH}, we predict an offset vector that links the center of each hand bounding box to the center of the corresponding active object bounding box. The offset vector is represented by a versor \textit{v} and a magnitude \textit{m}. This module is composed of a MLP with two fully connected layers. It takes as input a ROI-pooled feature vector extracted from the enlarged hand crop and infers the \textless\textit{$v_x, v_y$, m\textgreater} triplet, where \textit{$v_x$} and \textit{$v_y$} represent the components of the versor \textit{v}. \\
\noindent\textbf{Matching algorithm:} The last component of the proposed system is a matching algorithm that takes as input the outputs from the previous modules to predict the \textit{\textless hand, contact\_state, active\_object, \textless other\_objects\textgreater\textgreater} quadruplet. The algorithm computes for each hand in contact with an object an image point ($p_{interaction}$) using the coordinates of the center of the hand bounding box and the corresponding offset vector. This point represents the predicted center of the active object bounding box. The active object is selected considering the object bounding box whose center is closest to the inferred $p_{interaction}$ point and also checking if the bounding box has a nonzero intersection with the bounding box of the considered hand.
%\noindent\textbf{Training details:} To train the network, we used the standard Faster R-CNN losses. In addition, for the \textit{hand side classification} and the \textit{hand state classification} modules, we used the standard binary cross-entropy loss, whereas for the \textit{offset vector regression module}, we used the mean squared error loss. The final loss is the sum of all the losses. Additional implementation details are provided in the supplementary material. 
%\begin{equation}
%Loss = L_{faster\_rcnn} + L_{side} + L_{state} + L_{vector}
%\end{equation}
\begin{table}[t]
	\begin{minipage}{0.47\linewidth}
		\centering
        \caption{Statistics of the three splits: Train, Validation and Test.}
        \label{tab_statistics_real_dataset_train_val_test}
        \resizebox{0.85\textwidth}{!}{%
            \begin{tabular}{lccc}
            \textbf{Split} & \textbf{Train} & \textbf{Val} & \textbf{Test} \\ \hline
            \textbf{\#Videos}   & 2 & 2 & 4 \\ 
            \textbf{\#images}   & 992 & 734 & 1,330 \\ 
            \textbf{\%images}   & 32.46 & 24.01 & 43.53 \\ 
            \textbf{\#Hands}    & 1,653 & 1,036 & 1,814 \\ 
            \textbf{\#Objects}  & 6,483 & 4,337 & 6,778 \\
            \textbf{\#Active Objects} & 1,090 & 662 & 1,120 \\ \hline
            \end{tabular} }
	\end{minipage} \hfill
	\begin{minipage}{0.47\linewidth}
        \centering
        \caption{Object detection results using different amounts of real data.}\label{tab_object_detection}
        \resizebox{0.85\textwidth}{!}{%
        \begin{tabular}{c|c|c}
            \textbf{Pretraining} & \textbf{Real Data\%} & \textbf{mAP\%} \\ \hline
            Synthetic  & 0   & 66.44 \\
            - & 10  & 53.27 \\
            Synthetic  & 10  & 72.69 \\
            - & 25  & 52.34 \\
            Synthetic  & 25  & 76.19 \\
            - & 50  & 71.17 \\
            Synthetic  & 50  & \textbf{77.29} \\
            - & 100 & 70.84 \\
            Synthetic  & 100 & \underline{77.14} \\ \hline
        \end{tabular} }
	\end{minipage}
\end{table}

\section{Experiments and Results}\label{sec_experiments_results}
%We performed several experiments that aim to: 1) understand the usefulness of using the synthetic data, together with a variable amount of real data (0\%, 10\%, 25\%, 50\%, 100\%), when the proposed method is tested on real data, 2) evaluate the performance of the proposed method for the EHOIs detection and active objects recognition in an industrial domain, and 3) compare the obtained results with different baseline approaches based on the state-of-the-art method of~\cite{Shan2020UnderstandingHH}.
We split the real dataset into training, validation, and test sets. Table~\ref{tab_statistics_real_dataset_train_val_test} reports statistics about these splits. We trained our models in two stages. In the first stage, the models have been trained using only synthetic data (i.e., 0\% of real data). In the second stage, we finetuned the models considering different amount of the real training data, namely, 10\%, 25\%, 50\%, and 100\%.
\subsection{Object Detection Performance}
We evaluated the object detection performance of our method considering 19 object categories. We used the mean Average Precision metric, with an \textit{Intersection over Union (IoU)} threshold of \textit{$0.5$} (\textit{mAP@50})\footnote{We used the following implementation: \url{https://github.com/cocodataset/cocoapi}}. We report the results in Table \ref{tab_object_detection}. The \textit{``Pretraining''} column indicates whether synthetic data were used to pretrain the models. The \textit{``Real Data\%''} column reports the percentage of real data used to finetune the models. The table shows the best results in bold, whereas the second best results are underlined. The results show that using only synthetic data to train the model (first row of Table \ref{tab_object_detection}) allows to achieve reasonable performance for this task (\textit{mAP} of $66.44\%$). The best result (\textit{mAP} of $77.29\%$) was obtained by the model pretrained on the synthetic dataset and finetuned with $50\%$ of the real dataset, while the second best result (\textit{mAP} of $77.14\%$) comes from the model pretrained on the synthetic dataset and finetuned with $100\%$ of the real dataset. The results also highlight how combining synthetic and real data allows to increase the performance for the object detection task. Indeed, all the models which have been pretrained using synthetic data outperformed the corresponding models trained only with real data, especially when little real data is available. Furthermore, it is worth noting that the model pretrained on the synthetic dataset and finetuned with 10\% of the real dataset obtained a higher performance (\textit{mAP} of $72.69\%$) than all the models trained using only the real data (\textit{mAP} of $70.84\%$ using 100\% of real data), which supports the usefulness of synthetic data. See supplementary material for qualitative results.  %Figure~\ref{fig_qualitative_examples_proposed} reports some qualitative examples obtained using the model pretrained on the synthetic dataset and finetuned with $100\%$ of the real data.

\begin{table}[t]
\centering
\caption{Results for the EHOI detection task.}\label{tab_portion_synth_dataset}
\resizebox{0.85\linewidth}{!}{
\begin{tabular}{c|c|c|c|c|c|c|c}
    \textbf{Pretraining} & \textbf{Real Data\%}  & \textbf{AP Hand} & \textbf{mAP Obj} & \textbf{AP H+Side} & \textbf{AP H+State} & \textbf{mAP H+Obj} & \textbf{mAP All} \\ \hline
    Synthetic   & 0   & 80.89 & 29.52 & 78.65 & 36.16 & 26.29 & 23.78 \\
    -           & 10  & 90.48 & 23.26 & 79.46 & \underline{50.44} & 21.79 & 18.59 \\
    Synthetic   & 10  & 81.69 & 34.19 & 80.28 & 48.59 & 30.98 & 28.14 \\
    -           & 25  & 90.46 & 18.83 & 80.28 & 49.25 & 17.50 & 15.92 \\
    Synthetic   & 25  & \underline{90.61} & 31.17 & 80.50 & 48.90 & 28.38 & 26.60 \\
    -           & 50  & 90.38 & 27.08 & 79.98 & 48.95 & 25.54 & 23.27 \\
    Synthetic   & 50  & \underline{90.61} & \textbf{36.23} & 79.69 & 48.81 & \underline{31.87} & \underline{30.50} \\
    -           & 100 & 90.47 & 26.29 & \underline{89.20} & 50.13 & 25.04 & 22.70 \\
    Synthetic   & 100 & \textbf{90.67} & \underline{35.43} & \textbf{89.37} & \textbf{50.58} & \textbf{34.09} & \textbf{32.61} \\ \hline
    \end{tabular}%
}
\end{table}

\subsection{Egocentric Human Object Interaction Detection}
%In this section, we report the experiments for the EHOIs detection task. In particular, the goal of these experiments is to establish the usefulness of the synthetic dataset for this task. To this end, we compared different models trained using synthetic data and different amounts of the real dataset. \\ 
We evaluated our method considering the following metrics: 1) \textit{AP Hand}: Average Precision of the hand detections; 2) \textit{mAP Obj}: mean Average Precision of the active objects; 3) \textit{AP H+Side}: Average precision of the hand detections when the correctness of the side (\textit{Left/Right}) is required; 4) \textit{AP H+State}: Average precision of the hand detections when the correctness of the contact state (\textit{In contact/No contact}) is required; 5) \textit{mAP H+Obj}: mean Average Precision of the active objects when the correctness of the associated hand is required, and 6) \textit{mAP All}: mean Average Precision of the hand detections when the correctness of the side, contact state, and associated active object are required. Note that, while most of these metrics are based on~\cite{Shan2020UnderstandingHH}, we modified the metrics influenced by active objects (i.e., \textit{mAP Obj}, \textit{mAP H+Obj}, and \textit{mAP All}) to include the recognition of the object categories (switching from AP to mAP). The results summarized in Table~\ref{tab_portion_synth_dataset} highlight that using synthetic data allows to achieve the best performance. Indeed, the model pretrained with synthetic data and finetuned with 100\% of the real dataset (last row) obtained the best results considering all the evaluation measures, except for the mAP Obj measure, in which it obtains the second best result of $35.43\%$. In particular, considering the mAP all measure ($32.61\%$), it outperforms the model trained using 100\% of real data ($22.70\%$) by a significant margin of 9.91\%. The model trained using only synthetic data (first row) outperforms all the models using only real data with respect to the evaluation measures influenced by the active objects. Indeed, the aforementioned model obtains the best results with respect to mAP Obj ($29.52\%$), mAP H+Obj ($26.29\%$), and mAP All ($23.78\%$). These performance scores are higher as compared to those achieved by the model trained with 50\% of real data (i.e., $27.08\%$, $25.54\%$ and $23.27\%$). Nevertheless, for the measures related to the hands (i.e., AP Hand, AP H+Side and AP H+State), the discussed method achieves limited performance, probably due to the gap between the real and synthetic domains. Please see Figure~\ref{fig_desired_output}-left for a qualitative example of the proposed method.
%\begin{figure}[t]
%    \centering
%    \includegraphics[width=\textwidth]{images/Qualtitative_examples_ours.jpg}
%    \caption{Qualitative results of the proposed method pretrained with the synthetic dataset and finetuned with the 100\% of the real data.} %\label{fig_qualitative_examples_proposed}
%\end{figure}

We also compare the proposed method with different baselines based on the state-of-the-art method introduced in~\cite{Shan2020UnderstandingHH}, which was pretrained on the large-scale dataset 100DOH~\cite{Shan2020UnderstandingHH} which contains over 100K labeled frames of HOIs. To be able to compare the proposed method with~\cite{Shan2020UnderstandingHH}, we extend the former to recognize the class of active objects following to two different approaches. The first approach consists in training a Resnet-18 CNN~\cite{he2015deep} to classify image patches extracted from the active object detections. We trained the classifier with four different sets of data: 1) BS1: 19 videos, one per object class, in which only the considered object is observed. This provides a minimal training set that can be collected with a modest labeling effort; %\footnoteref{seesup}; %\footnote{See the supplementary material for additional details.}; 
2) BS2: images sampled from the proposed real dataset; 3) BS3: synthetic data and 4) BS4: both real and synthetic data. Note that this set requires a significant data collection and labeling effort. The second approach (BS5) uses the YOLOv5\footnote{YOLOv5: \url{https://github.com/ultralytics/yolov5}} object detector to assign a label to the active objects predicted by~\cite{Shan2020UnderstandingHH}. In particular, to each active object, we assign the class of the object with the highest IoU among those predicted by YOLOv5, otherwise, if there are no box intersections, the proposal is discarded. Table~\ref{tab_comparison_hic} reports the obtained results. Considering the measures based on the active objects (i.e., mAP Obj, mAP H+Obj, and mAP all), our method trained with 50\% of the real data (496 images) outperforms all the baselines based on~\cite{Shan2020UnderstandingHH} obtaining performances of $36.23\%$, $31.87\%$ and $30.50\%$ respectively. Moreover, using 100\% of the real data, our approach obtains comparable performance considering the measures based on the hands (i.e., AP Hand, AP H+Side, and AP H+State). See supplementary material for qualitative comparison. %is shown in Figure \ref{fig_comparative_examples_hic_proposed}.
\begin{table}[t]
    \centering
    \caption{Comparison between the proposed method and different baseline approaches based on~\cite{Shan2020UnderstandingHH}.}
    \label{tab_comparison_hic}
    \resizebox{0.9\linewidth}{!}{%
    \begin{tabular}{c|c|c|c|c|c|c|c|c}
    \textbf{} & \textbf{Pretraining} & \textbf{Real Data\%}  & \textbf{AP Hand} & \textbf{mAP Obj} & \textbf{AP H+Side} & \textbf{AP H+State} & \textbf{mAP H+Obj} & \textbf{mAP All} \\ \hline
    \textbf{Proposed method} & Synthetic & 0            & 80.89 & 29.52 & 78.65 & 36.16 & 26.29 & 23.78 \\
    \textbf{Proposed method} & Synthetic & 50    & 90.61 & \textbf{36.23} & 79.69 & 48.81 & \underline{31.87} & \underline{30.50} \\
    \textbf{Proposed method} & Synthetic & 100   & \underline{90.67} & \underline{35.43} & \underline{89.37} & \underline{50.58} & \textbf{34.09} & \textbf{32.61} \\ \hline\hline
    \multicolumn{9}{c}{} \\ \hline
    \textbf{BS1} & - & 100 & \textbf{90.76} & 14.10 & \textbf{89.78} & \textbf{59.23} & 12.42 & 11.23 \\
    \textbf{BS2} & - & 100                & \textbf{90.76} & 22.17 & \textbf{89.78} & \textbf{59.23} & 19.76 & 18.05 \\
    \textbf{BS3} & Synthetic & 0             & \textbf{90.76} & 09.44 & \textbf{89.78} & \textbf{59.23} & 08.51 & 07.49  \\
    \textbf{BS4} & Synthetic & 100    & \textbf{90.76} & 26.36 & \textbf{89.78} & \textbf{59.23} & 21.00 & 19.48 \\
    \textbf{BS5} & Synthetic & 100    & \textbf{90.76} & 30.26 & \textbf{89.78} & \textbf{59.23} & 28.77 & 27.12 \\ \hline
    \end{tabular}%
    }
\end{table}
\begin{comment}
\begin{figure}[t]
  \centering
  \begin{subfigure}[b]{0.40\linewidth}
    \includegraphics[width=\linewidth]{images/examples/53_48074_our.jpg}
    \caption{Proposed system}
  \end{subfigure}
  \begin{subfigure}[b]{0.40\linewidth}
    \includegraphics[width=\linewidth]{images/examples/53_48074_hic.jpg}
    \caption{Method of~\cite{Shan2020UnderstandingHH} (BS5)}
  \end{subfigure}
  \caption{Comparison between our method trained with synthetic data and 100\% of the real dataset (a) and BS5 based on~\cite{Shan2020UnderstandingHH} (b). In (b) the left hand is incorrectly associated to the screwdriver.}
  \label{fig_comparative_examples_hic_proposed}
\end{figure}
\end{comment}

\section{Conclusion}\label{sec_conclusion}
We considered the EHOI detection task in the industrial domain. Since labeling images is expensive in terms of time and costs, we explored how using synthetic data can improve the performance of EHOIs detection systems. To this end, we generated a new dataset of automatically labeled photo-realistic synthetic EHOIs in an industrial scenario  and collected 8 egocentric real videos, which have been manually labeled. We proposed a method to tackle EHOI detection and compared it with different baseline approaches based on the state-of-the-art method of~\cite{Shan2020UnderstandingHH}.
%that is able to detect and recognize the hands and the objects and infer different information, such as: 1) active/no active object categories, 2) hand side, 3) hand contact state and 4) EHOIs as \textit{\textless hand, contact state, active object\textgreater} triplet. Furthermore, we compared our method with different baseline approaches based on the state-of-the-art method of~\cite{Shan2020UnderstandingHH}. 
Our analysis shows that exploiting synthetic data to train the proposed method greatly improves performance when tested on real data. Future work will explore how the knowledge of active/no active objects, inferred by our system, can provide useful information for other tasks such as next active object prediction.
\section*{Acknowledgements}
This research has been supported by Next Vision\footnote{Next Vision: \url{https://www.nextvisionlab.it/}} s.r.l., by the project MISE - PON I\&C 2014-2020 - Progetto ENIGMA - Prog n. F/190050/02/X44 – CUP: B61B19000520008, and by Research Program Pia.ce.ri. 2020/2022 Linea 2 - University of Catania.

\section*{\uppercase{Supplementary Material}}

\section{Additional Details on the Dataset}\label{sec_dataset_sup}
\begin{figure}[t]
    \centering
    \includegraphics[width=\textwidth]{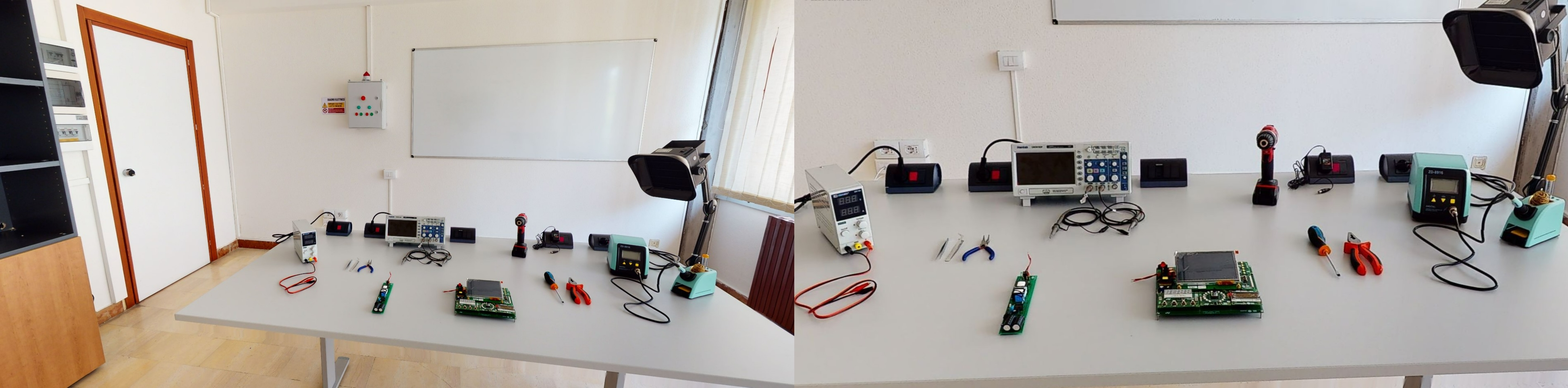}
    \caption{Industrial-like laboratory.}
    \label{fig_industrial_laboratory}
\end{figure}

\begin{figure}[t]
    \centering
    \includegraphics[width=\textwidth]{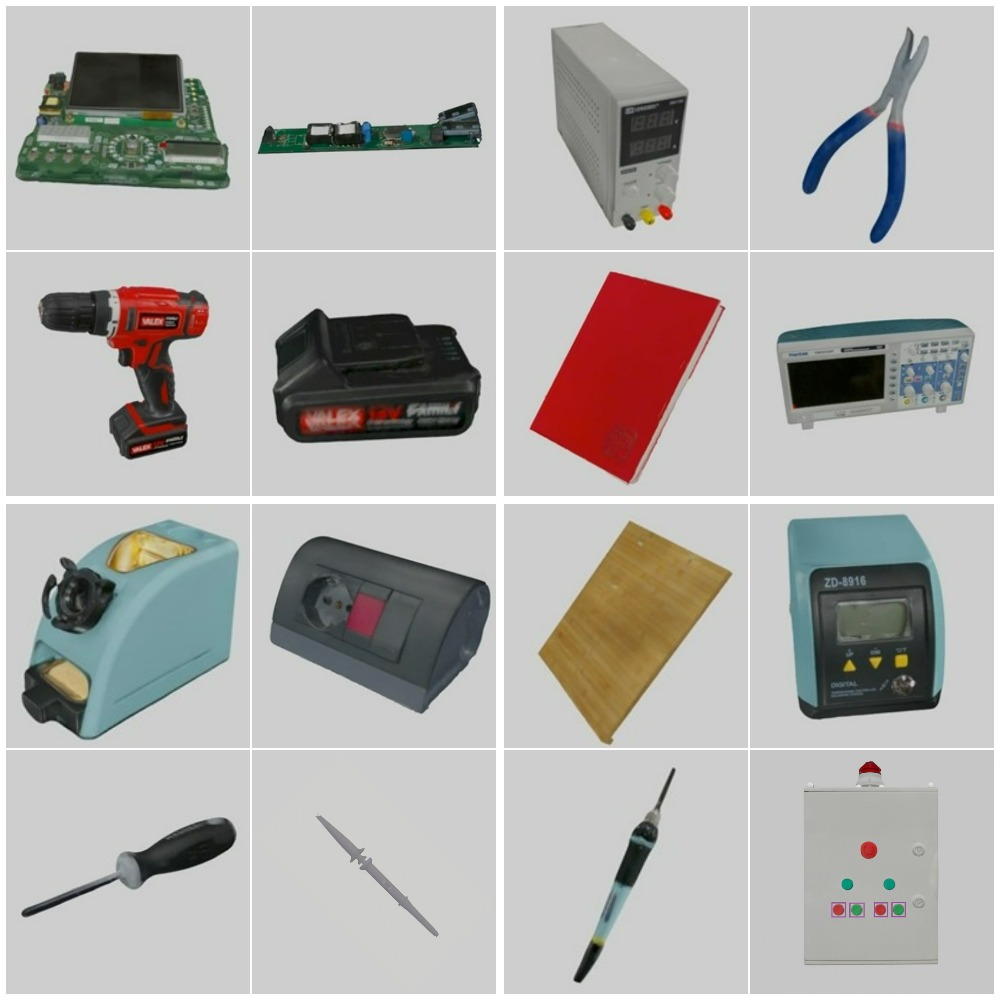}
    \caption{Objects categories.}
    \label{fig_objects_categories}
\end{figure}

\begin{figure}[t]
    \centering
    \begin{subfigure}[b]{0.45\linewidth}
        \centering
        \includegraphics[width=1\textwidth]{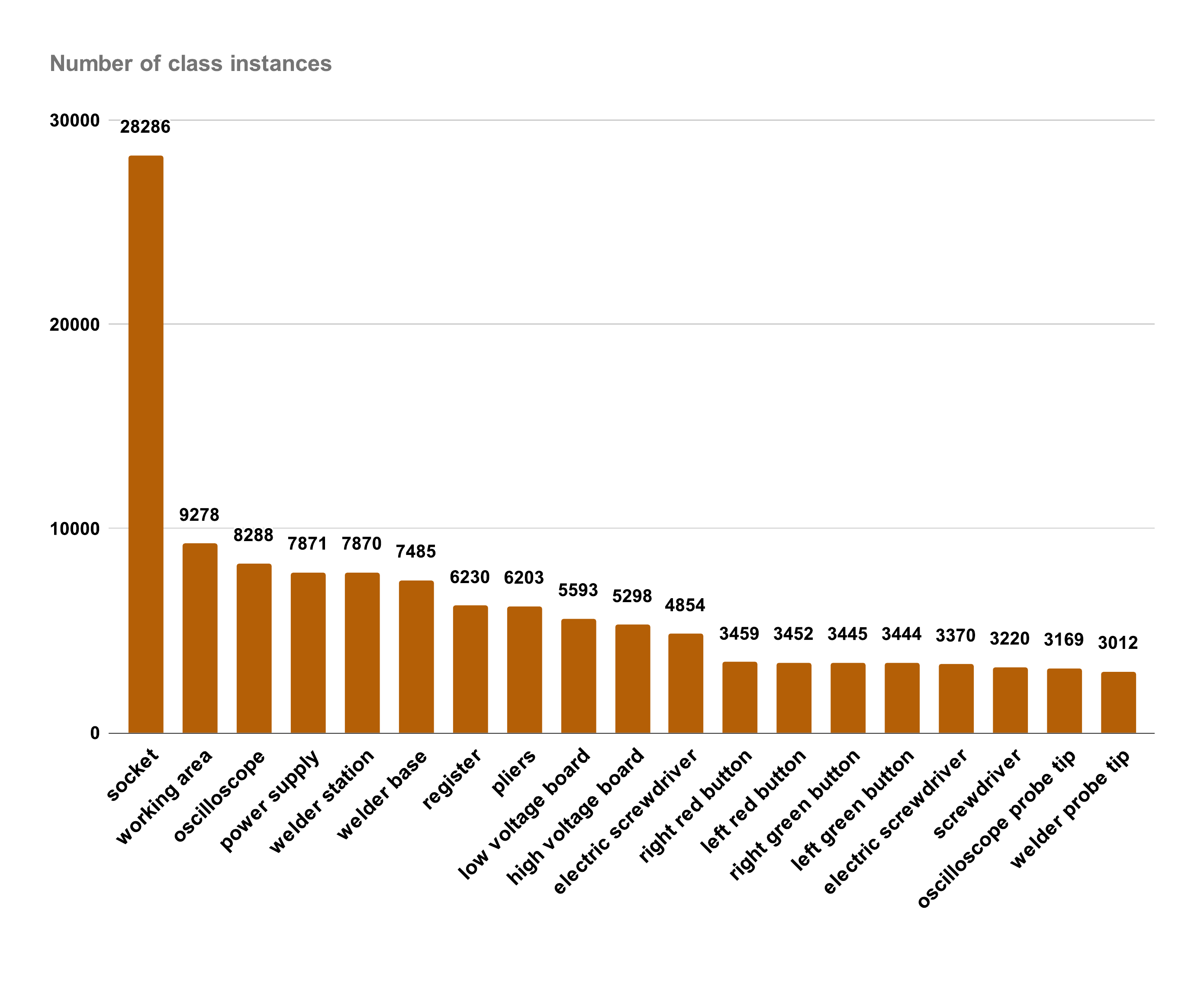}
        \caption{Distributions of the all objects instances in the synthetic dataset.}
    \end{subfigure}
    \begin{subfigure}[b]{0.45\linewidth}
        \centering
        \includegraphics[width=1\textwidth]{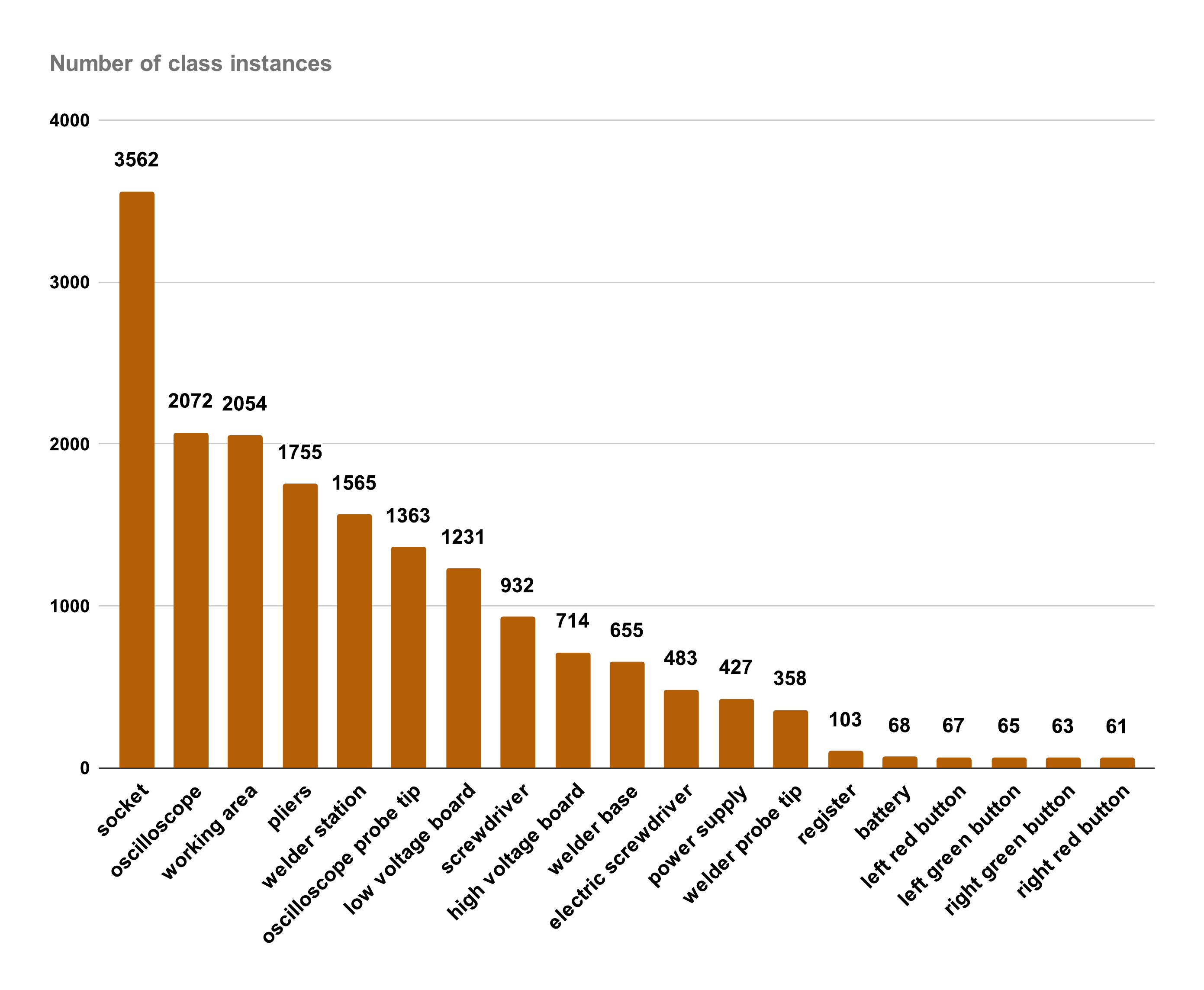}
        \caption{Distributions of the all objects instances in the real dataset.}
    \end{subfigure}
    \begin{subfigure}[b]{0.45\linewidth}
        \centering
        \includegraphics[width=1\textwidth]{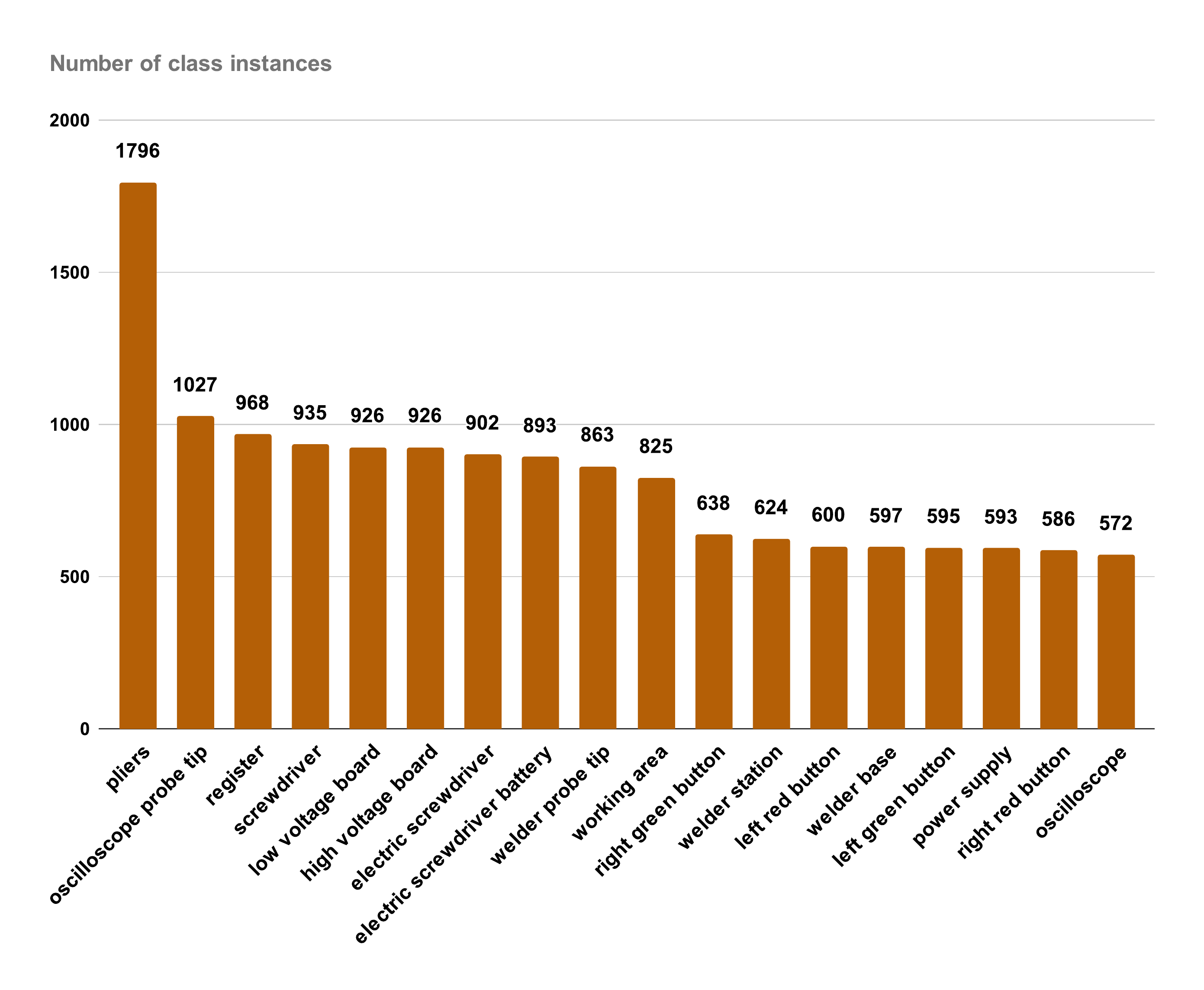}
        \caption{Distributions of the active objects instances in the synthetic dataset.}
    \end{subfigure}
    \begin{subfigure}[b]{0.45\linewidth}
        \centering
        \includegraphics[width=1\textwidth]{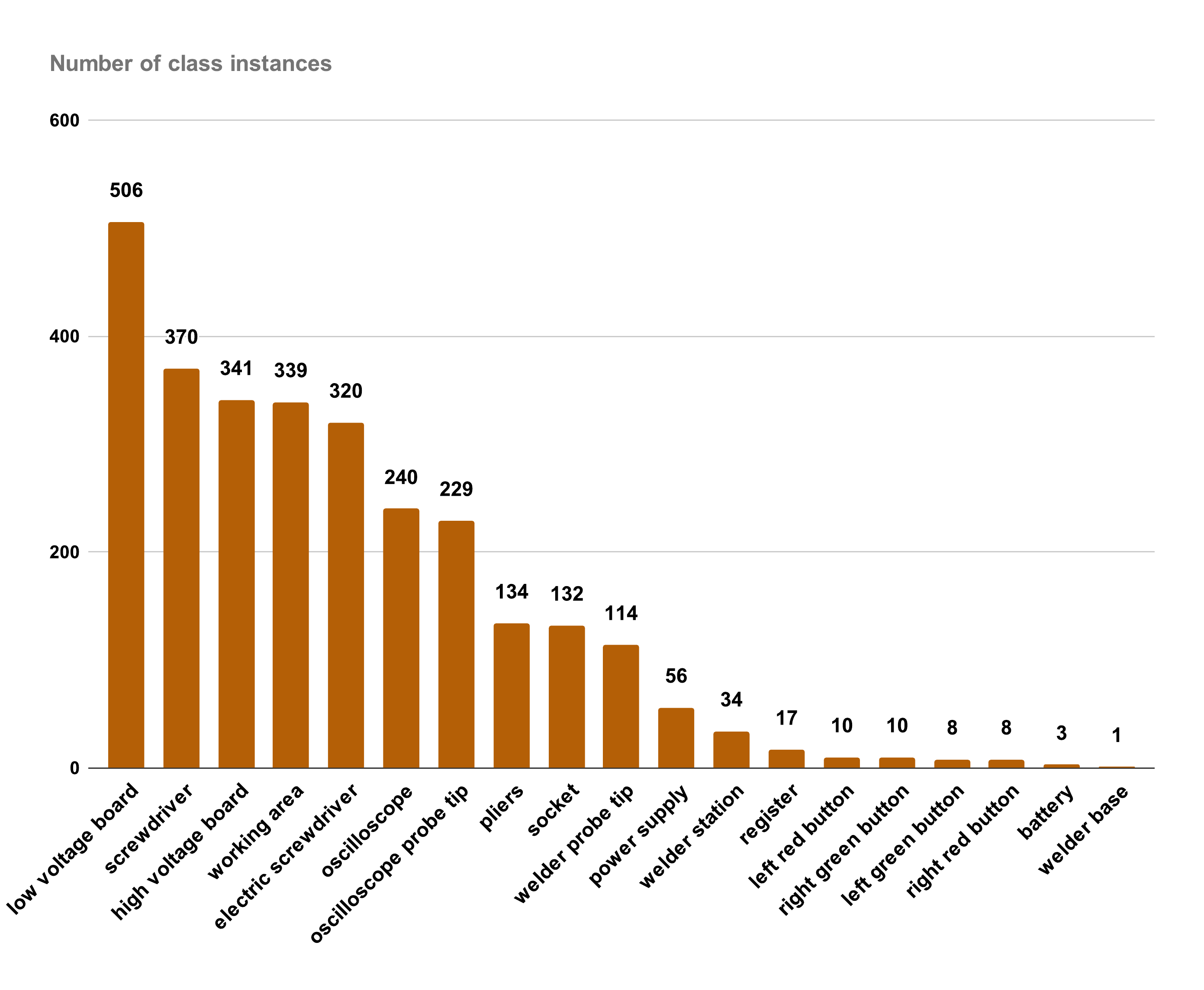}
        \caption{Distributions of the active objects instances in the real dataset.}
    \end{subfigure}
    \caption{The distributions of all/active objects instances in the two datasets.}
    \label{fig_distributions_instances}
\end{figure}

\paragraph{Context:} Figure \ref{fig_industrial_laboratory} shows the laboratory that we set up to study the EHOIs detection task in industrial domain. 
\paragraph{Categories:} We considered the following 19 objects categories: \textit{power supply, oscilloscope, welder station, electric screwdriver, screwdriver, pliers, welder probe tip, oscilloscope probe tip, low voltage board, high voltage board, register, electric screwdriver battery, working area, welder base, socket, left red button, left green button, right red button} and \textit{right green button}. Figure \ref{fig_objects_categories} shows all the objects categories.
\paragraph{Privacy:} During the acquisitions of the real dataset, the subjects removed all the items that could have revealed their identity (e.g., rings, watches, etc.). No other subjects appear in the captured videos.
\paragraph{Statistics:} Table \ref{tab_statistics_dataset_synthetic} reports the statistics related to the synthetic generated dataset, while Table \ref{tab_statistics_real_dataset} reports the statistics of the real dataset. Figure \ref{fig_distributions_instances} shows the distributions of all/active objects instances of the synthetic and the real datasets. \\
\begin{table}[t]
	\begin{minipage}{0.45\linewidth}
		\centering
        \caption{Statistics of the generated synthetic dataset.}
        \label{tab_statistics_dataset_synthetic}
        \resizebox{\linewidth}{!}{%
        \begin{tabular}{l|c}
            \textbf{Total number of images} & 20,000    \\ \hline
            \textbf{\#hands}                & 29,034    \\
            \textbf{\#hands in contact}     & 14,589    \\
            \textbf{\#hands not in contact} & 14,445    \\
            \textbf{\#left hands}           & 14,473    \\
            \textbf{\#right hands}          & 14,561    \\
            \textbf{\#object categories}    & 19        \\ 
            \textbf{\#objects}              & 123,827   \\
            \textbf{\#active objects}       & 14,589    \\ \hline
        \end{tabular} } 
	\end{minipage} \hfill
	\begin{minipage}{0.45\linewidth}
        \centering
        \caption{Statistics of the real dataset.}
        \label{tab_statistics_real_dataset}
        \resizebox{\linewidth}{!}{%
        \begin{tabular}{l|c}
            \textbf{\#videos} & 8  \\ \hline
            \textbf{Total videos length} & 227 min \\
            \textbf{Avg. video duration} & 28.37 min \\
            \textbf{\#subjects} & 7  \\
            \textbf{\#images} & 3,056  \\
            \textbf{\#hands} & 4,503     \\
            \textbf{\#hands in contact} & 3,311 \\
            \textbf{\#hands not in contact} & 1,192  \\
            \textbf{\#left hands} & 2,013     \\
            \textbf{\#right hands} & 2,490  \\
            \textbf{\#object categories}  & 19 \\ 
            \textbf{\#objects} & 17,598 \\
            \textbf{\#active objects} & 2,872\\\hline
        \end{tabular} }
	\end{minipage}
\end{table}

\subsection{Additional Dataset}
\begin{figure}[t]
    \centering
    \includegraphics[width=\textwidth]{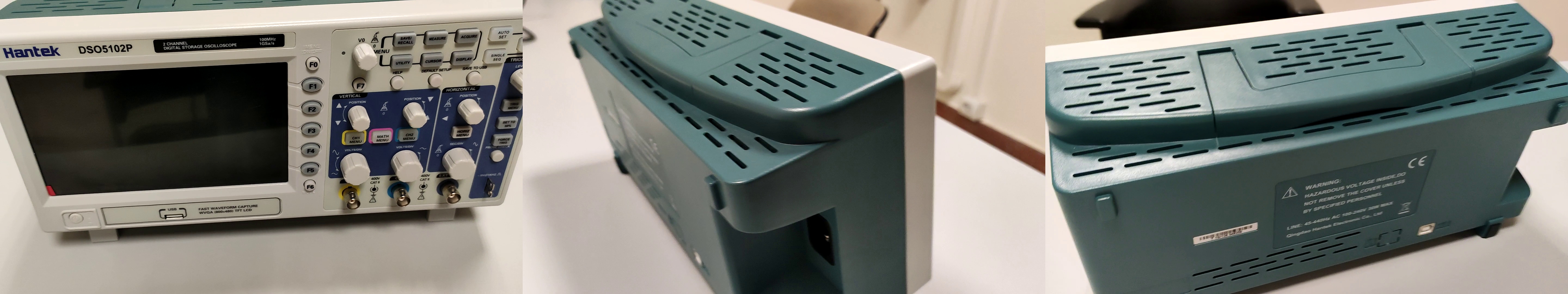}
    \caption{Examples of frames in which the object has been acquired from different points of view.}
    \label{fig_oscilloscope_views}
\end{figure}

In order to train the classifier used in the baseline \textit{BS1}, we collected and annotated an additional set of data in the same environment. To this end, we acquired 19 videos (i.e., one for each object class) at a resolution of 1920x1080 pixels and with a framerate of 30fps using a smartphone. Each video shows a single object from different points of view. From these videos, we have extracted about 27,000 frames labeled with the object category. Figure~\ref{fig_oscilloscope_views} shows some examples of the extracted frames.

\section{Details of the Proposed Approach}\label{sec_method_sup}
\subsection{Network Details}
\begin{figure}[!htb]
    \centering
    \begin{subfigure}[b]{0.35\linewidth}
        \centering
        \includegraphics[width=0.8\linewidth]{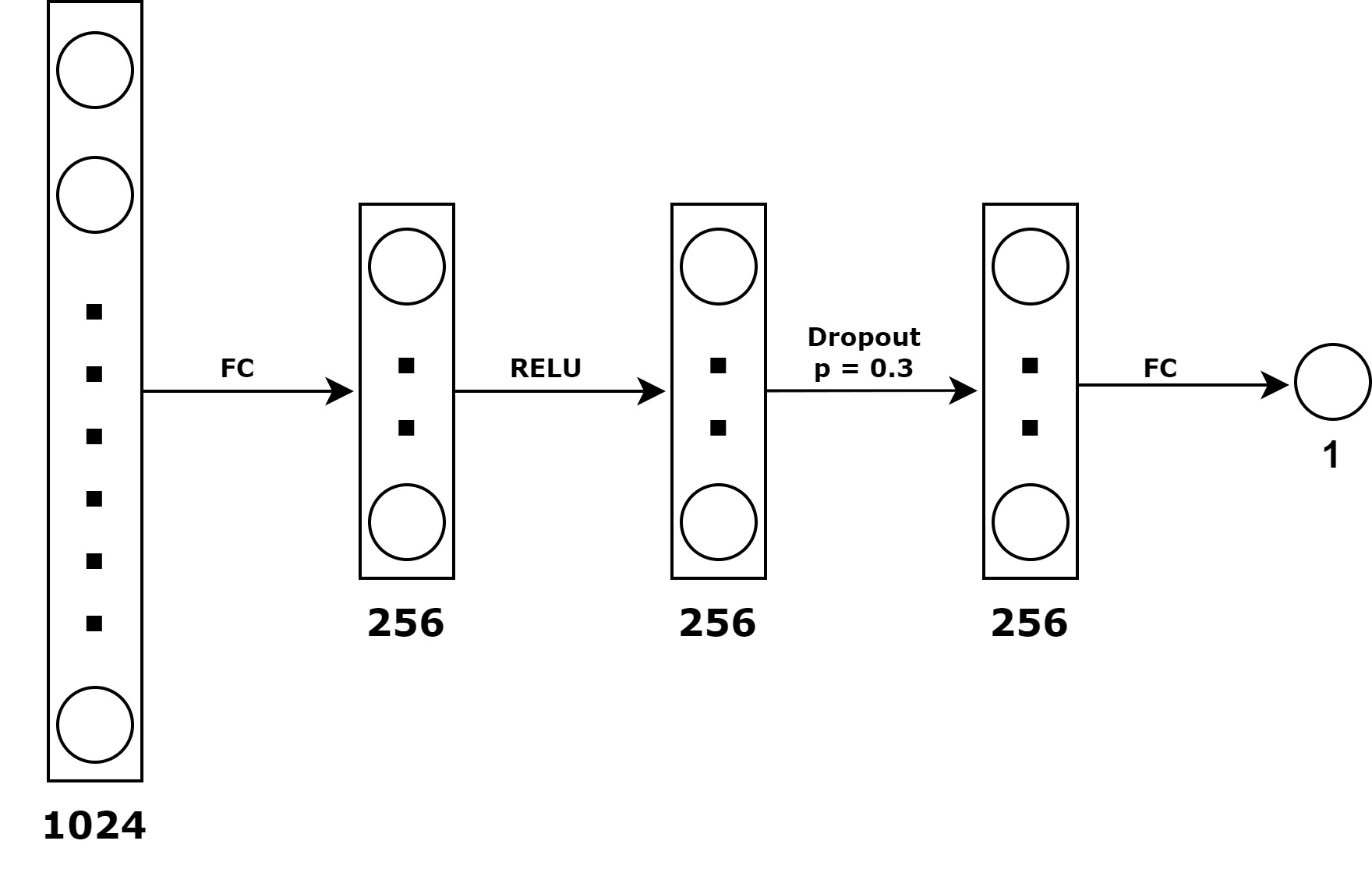}
        \caption{Hand side classification module.}
    \end{subfigure}
    \vspace{1em}
    \begin{subfigure}[b]{0.35\linewidth}
        \centering
        \includegraphics[width=0.8\linewidth]{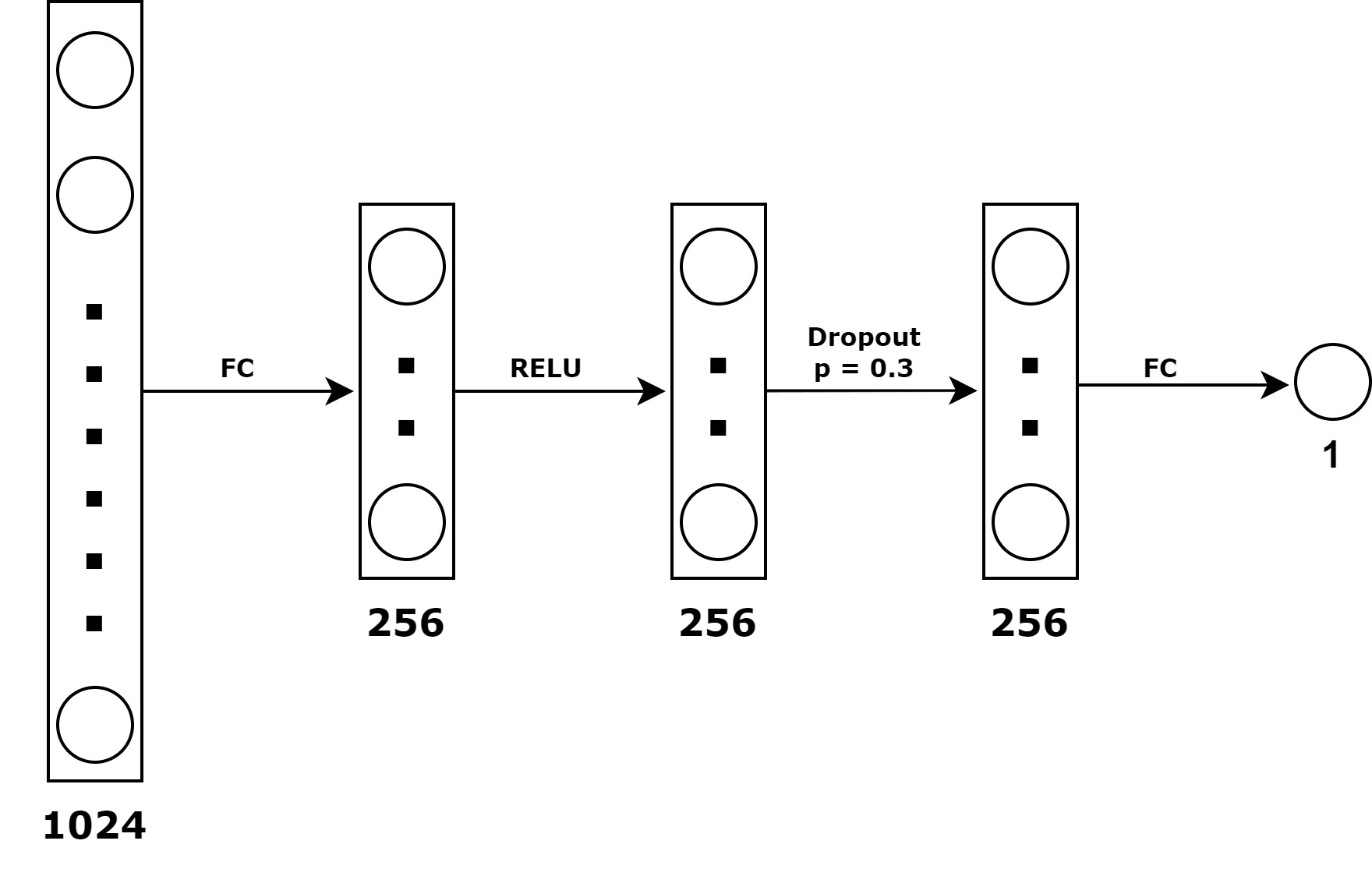}
        \caption{Hand state classification module.}
    \end{subfigure}
    \vspace{1em}
    \begin{subfigure}[b]{0.27\linewidth}
        \centering
        \includegraphics[width=0.8\linewidth]{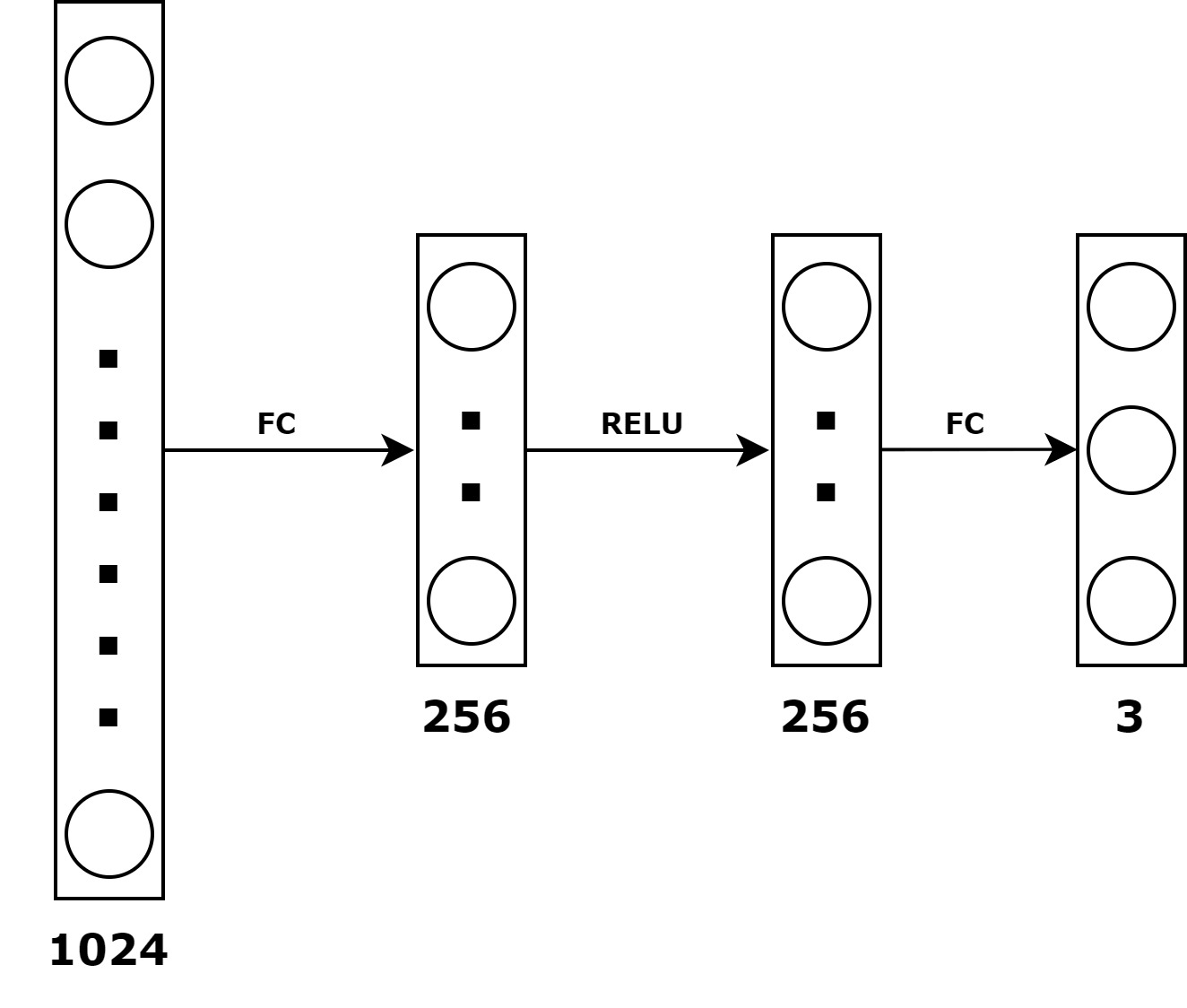}
        \caption{Offset vector regression module.}
    \end{subfigure}
    \caption{Architectures of the proposed modules.}
    \label{fig_modules}
\end{figure}

We built our system by extending the implementation of Faster R-CNN~\cite{faster_rccn} from the detectron2 framework\footnote{\url{https://github.com/facebookresearch/detectron2}} with four additional components: 1) the hand side classification module, 2) the hand state classification module, 3) the offset vector regression module, and 4) the matching algorithm. Figure~\ref{fig_modules} shows the architectures of the modules.

\subsection{Synthetic Motion Blur}
\begin{figure}[t]
    \centering
    \includegraphics[width=.7\linewidth]{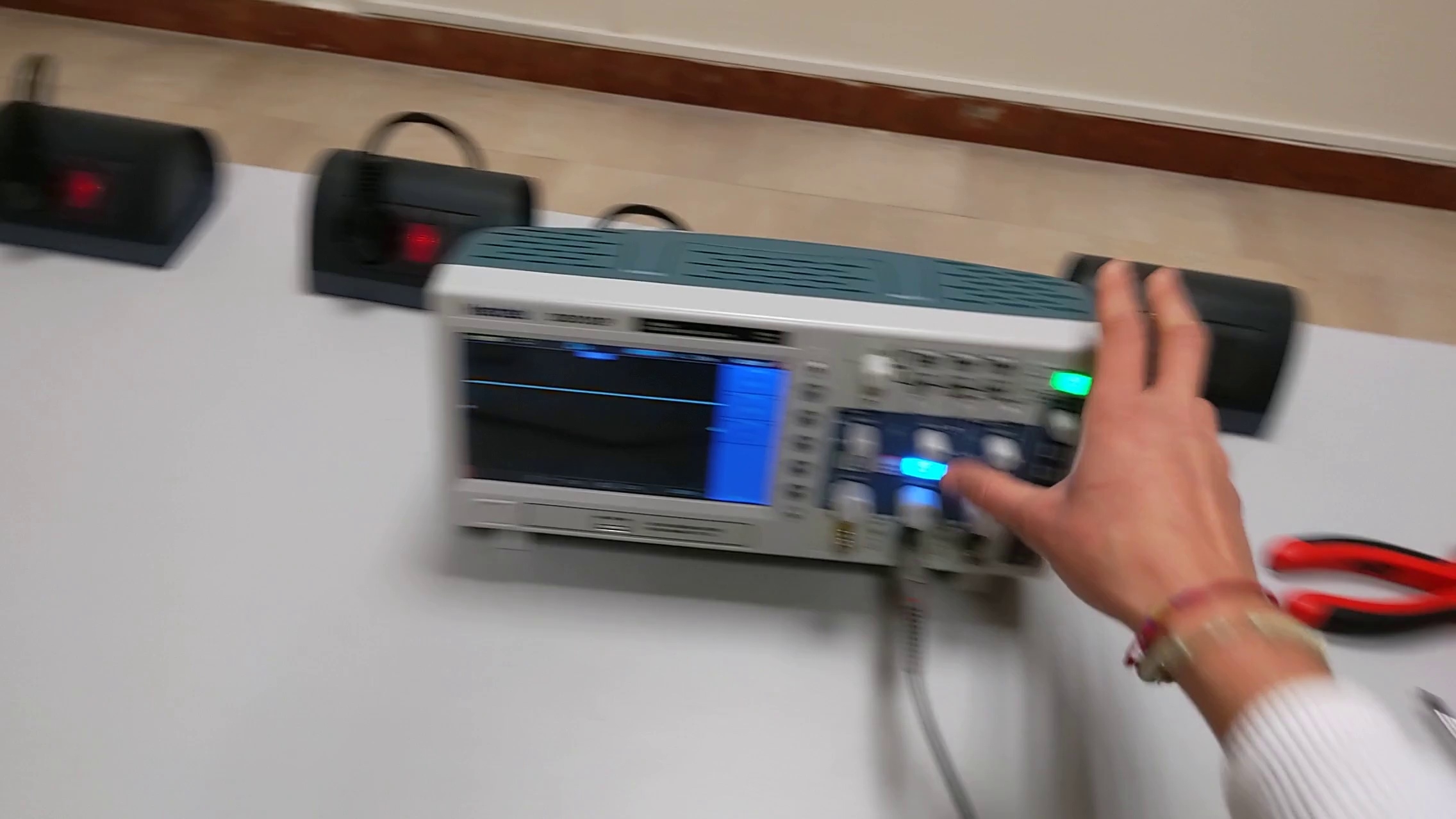}
    \caption{Example of blurred frame due to camera motion.}
    \label{fig_real_motion_blur}
\end{figure}

\begin{figure}[t]
    \centering
    \begin{subfigure}[b]{0.31\linewidth}
        \includegraphics[width=\linewidth]{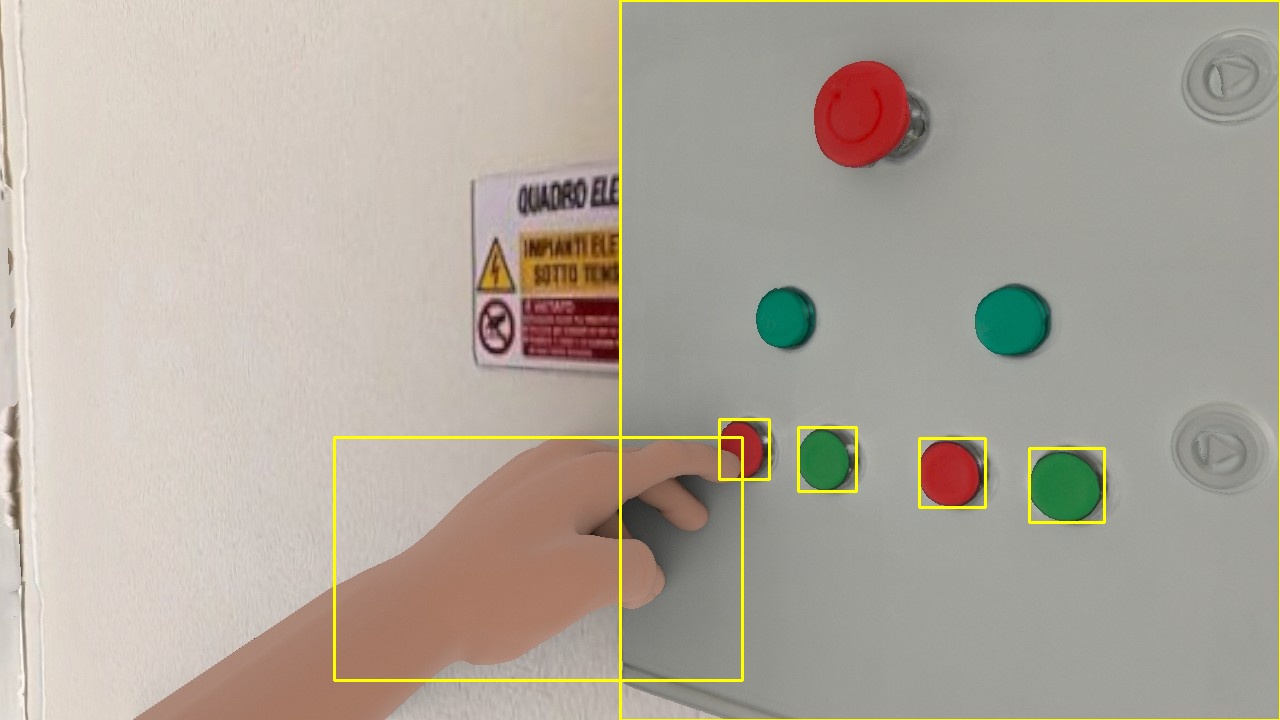}
        \caption{Synthetic image}
    \end{subfigure}
    \begin{subfigure}[b]{0.31\linewidth}
        \includegraphics[width=\linewidth]{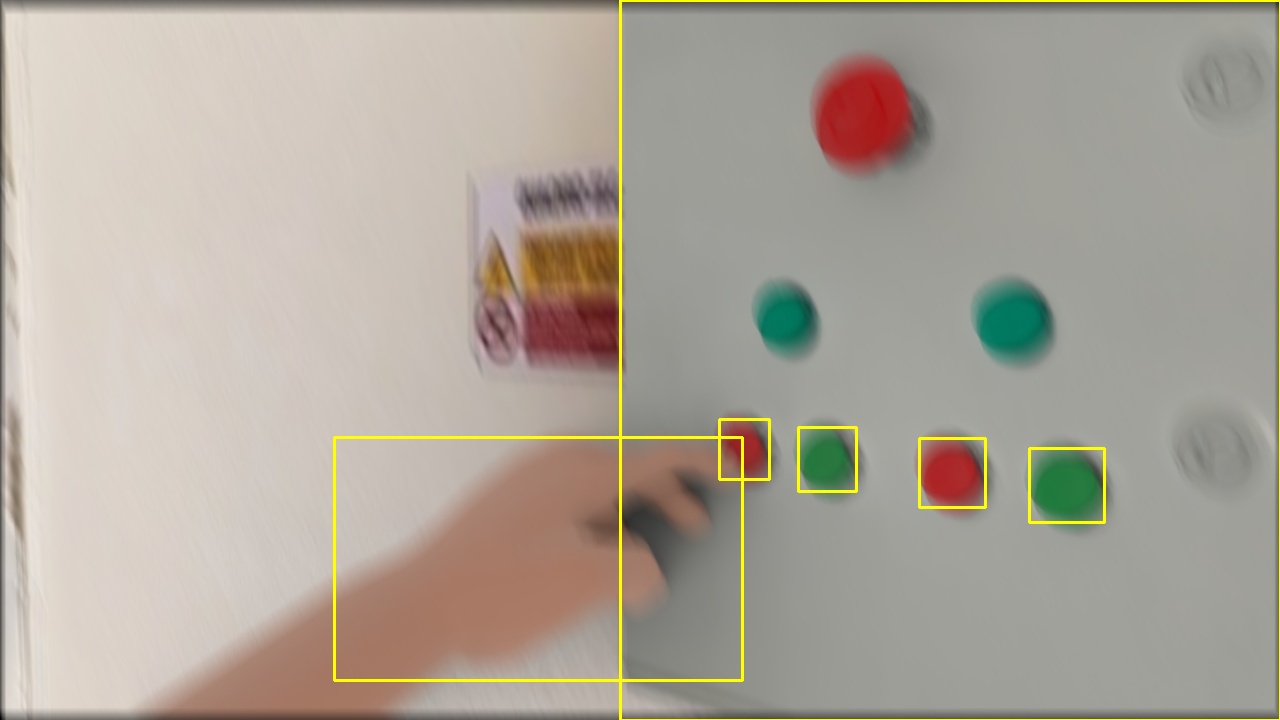}
        \caption{Blurred image}
    \end{subfigure}
    \begin{subfigure}[b]{0.31\linewidth}
        \includegraphics[width=\linewidth]{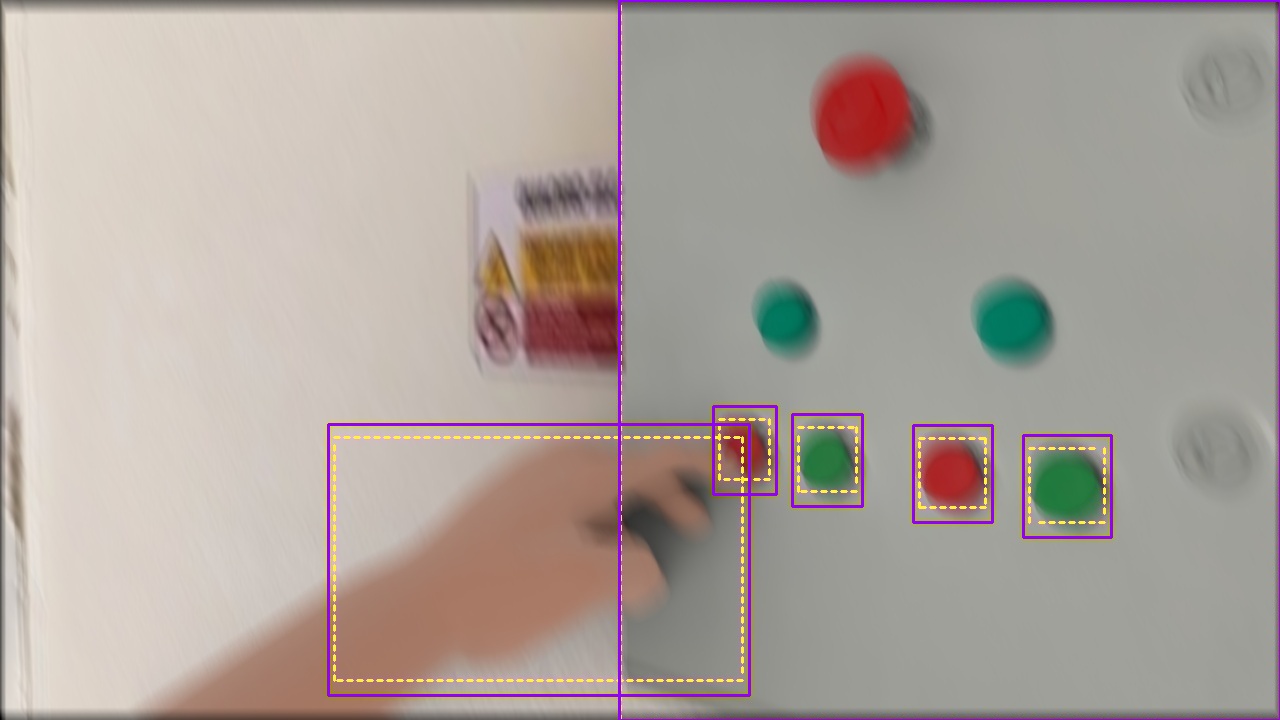}
        \caption{Boxes correction}
    \end{subfigure}
    \caption{Figure (a) shows the image before applying the synthetic motion blur, whereas in (b) is shown the same image after the application of motion blur, and (c) shows the correction process of the bounding boxes (indicated with purple boxes).}
    \label{fig_synthetic_motion_blur_w_box}
\end{figure}

Since in the real dataset several frames are blurred due to the camera motion (see Figure~\ref{fig_real_motion_blur}), we adopted a non-linear motion blur procedure~\cite{Sayed2021ImprovedHO} on the synthetic images to further reduce the gap between real and synthetic domains. \\
\noindent During the training phase, given a synthetic image (Figure~\ref{fig_synthetic_motion_blur_w_box} - (a)) we applied a non-linear motion blur kernel (Figure~\ref{fig_synthetic_motion_blur_w_box} - (b)). As shown in the work of~\cite{Sayed2021ImprovedHO}, correcting the coordinates of the bounding boxes further increases object detection performance. To do this, we applied the same kernel to the related semantic mask obtaining the new coordinates of the objects bounding boxes (Figure~\ref{fig_synthetic_motion_blur_w_box} - (c)).

\section{Additional Experiments and Results}\label{sec_experiments_results_sup}
\subsection{Training Details}
We used an Nvidia V100 GPU to perform the experiments. To train the model with the synthetic dataset, we initialized the learning rate to 0.001 and set a warm-up factor of 1000. We set the batch size to 4 and trained for 50,000 iterations. Instead, for training with real data, we decreased the learning rate by a factor of 10 after 12,500 and 15,000 iterations. We have set the batch size to 2 and trained for 20,000 iterations. Lastly, all the images were rescaled to $1280$x$720$ pixels for both the training and testing phases. \\
\indent To train the network, we used the standard Faster R-CNN losses. In addition, for the \textit{hand side classification} and the \textit{hand state classification} modules, we used the standard binary cross-entropy loss, whereas for the \textit{offset vector regression module}, we used the mean squared error loss. The final loss is the sum of all the losses:
\begin{equation}
Loss = L_{faster\_rcnn} + L_{side} + L_{state} + L_{vector}
\end{equation}
\subsection{Object Detection Performance}
\begin{figure}[t]
    \centering
    \begin{subfigure}[b]{0.47\linewidth}
        \centering
        \includegraphics[width=\textwidth]{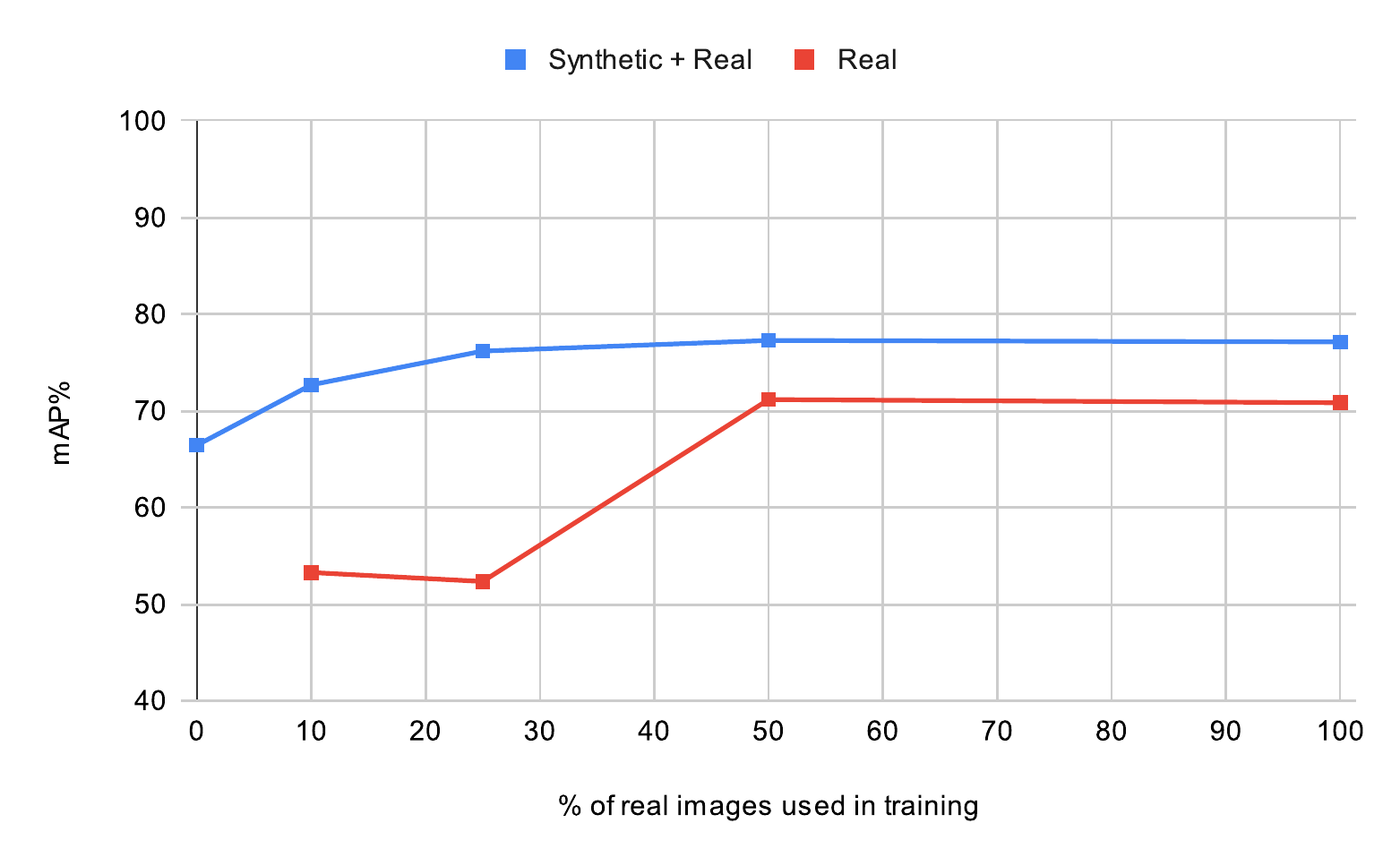}
        \caption{\textit{mAP}}
    \end{subfigure}
    \begin{subfigure}[b]{0.47\linewidth}
        \centering
        %\includepdf[width=\textwidth]{images/mAP_Hand-ALL.pdf}
        \includegraphics[width=\textwidth]{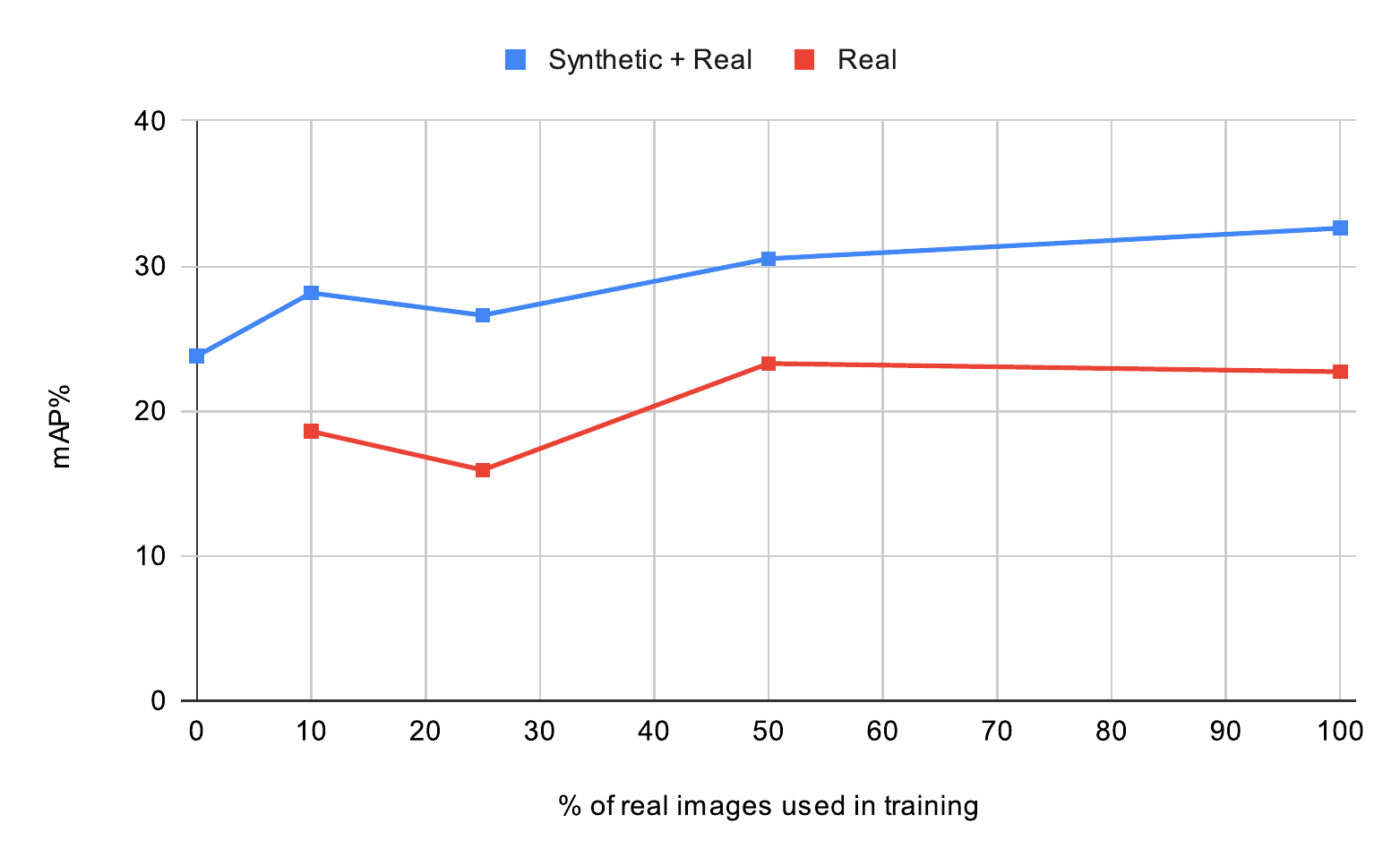}
        \caption{\textit{mAP All}}
    \end{subfigure}
    \caption{Comparison of the results of the proposed method on the real test data in term of ``\textit{mAP}'' (a) and ``\textit{mAP All}'' (b). The blue curves report the results of the models pretrained using synthetic data and finetuned with different amount of real data, while the red curves report the results of the models that used only real data.}
    \label{fig_graphs}
\end{figure}

Figure~\ref{fig_graphs} - (a) shows a comparison of the object detection performance, considering the ``\textit{mAP}'' curves, between the models
trained using synthetic data and different amount of real data (0\%, 10\%, 25\%, 50\%, 100\%). Table~\ref{tab_results_ap_obj} reports the AP of each class. Note that most of the best results come from the models pretrained using the synthetic data.
\begin{table}[t]
\centering
\caption{Object detection results per class using different amounts of real data. Best results in bold.}
\label{tab_results_ap_obj}
\resizebox{0.95\textwidth}{!}{%
\begin{tabular}{l|c|c|c|c|c||c|c|c|c} 
    \hline
    \multicolumn{1}{l|}{\multirow{4}{*}{\textbf{Category}}} & \multicolumn{9}{c}{\textbf{AP\%}} \\ \cline{2-10}
    & \multicolumn{9}{c}{\textbf{Real Data\%}} \\ \cline{2-10}
    & \multicolumn{5}{c||}{\textbf{Pretraining Synthetic}} & \multicolumn{4}{c}{\textbf{No Pretraining}} \\ \cline{2-10}
    & \textbf{0\%} & \textbf{10\%} & \textbf{25\%} & \textbf{50\%} & \textbf{100\%} & \textbf{10\%} & \textbf{25\%} & \textbf{50\%} & \textbf{100\%} \\ \hline
    \textbf{power supply} & 77.34 & 78.48 & 79.92 & \textbf{88.51} & 70.24 & 74.48 & 58.69 & 70.32 & 59.8 \\
    \textbf{oscilloscope} & 80.64 & 89.59 & 89.27 & \textbf{90.36} & 90.17 & 80.35 & 80.79 & 81.03 & 89.74 \\
    \textbf{welder station} & 80.86 & \textbf{89.59} & 89.25 & 89.49 & 89.52 & 88 & 89.12 & 88.51 & \textbf{89.59} \\
    \textbf{electric screwdriver} & 75.38 & 71.68 & 71.32 & 80.04 & 79.98 & 62.2 & 70.93 & 71.62 & \textbf{80.78} \\
    \textbf{screwdriver} & 42.76 & 50.28 & 53.68 & 56.41 & 55.83 & 40.96 & 48.6 & 53.6 & \textbf{57.11} \\
    \textbf{pliers} & 78.78 & 79.83 & 79.36 & 80.37 & \textbf{80.77} & 77.77 & 78.29 & 71.1 & 80.58 \\
    \textbf{welder probe tip} & 9.09 & 32.64 & 47.87 & 49.5 & \textbf{50.37} & 20.29 & 37.6 & 38.67 & 47.7 \\
    \textbf{oscilloscope probe tip} & 1.82 & 40.01 & 41.41 & 38.64 & \textbf{49.93} & 37.85 & 40.98 & 42.52 & 43.28 \\
    \textbf{low voltage board} & 62.59 & 70.66 & 79.45 & 79.98 & \textbf{80.46} & 68.15 & 75.88 & 79.26 & 80.13 \\
    \textbf{high voltage board} & 41.45 & 51.44 & 59.08 & 51.28 & \textbf{61.49} & 50.59 & 49.81 & 54.84 & 57.93 \\
    \textbf{register} & 58.48 & 54.49 & 79.05 & 68.77 & 70.35 & 56.88 & 63.64 & \textbf{79.09} & 75.22 \\
    \textbf{electric screwdriver battery} & 68.87 & 53.9 & 58.33 & 61.16 & 62.12 & 18.18 & 31.17 & 58.6 & \textbf{62.81} \\
    \textbf{working area} & 69.23 & 79.5 & 79.03 & 79.67 & 79.68 & 77.76 & 77.54 & 77.8 & 78.77 \\
    \textbf{welder base} & 81.35 & 81.77 & 81.73 & \textbf{90.79} & 81.82 & 71.37 & 81.07 & 81.47 & 81.73 \\
    \textbf{socket} & 61.45 & 85.08 & 88.48 & 89.81 & 90.09 & 86.25 & 89.82 & 90.19 & \textbf{90.43} \\
    \textbf{left red button} & \textbf{90.91} & \textbf{90.91} & \textbf{90.91} & \textbf{90.91} & \textbf{90.91} & 9.09 & 0 & 72.73 & 54.55 \\
    \textbf{left green button} & 88.48 & 88.48 & 87.88 & \textbf{88.79} & 88.18 & 13.64 & 6.06 & 70.55 & 62.34 \\
    \textbf{right red button} & 95.82 & \textbf{96.06} & 94.55 & 95.83 & 95.3 & 51.13 & 14.55 & 86.57 & 67.74 \\
    \textbf{right green button} & 96.97 & 96.67 & 96.97 & 98.18 & \textbf{98.48} & 27.27 & 0 & 83.8 & 85.74 \\ \hline
\end{tabular}}
\end{table}

\subsection{Egocentric Human Object Interaction Detection}
\begin{figure}[t]
    \centering
    \includegraphics[width=\textwidth]{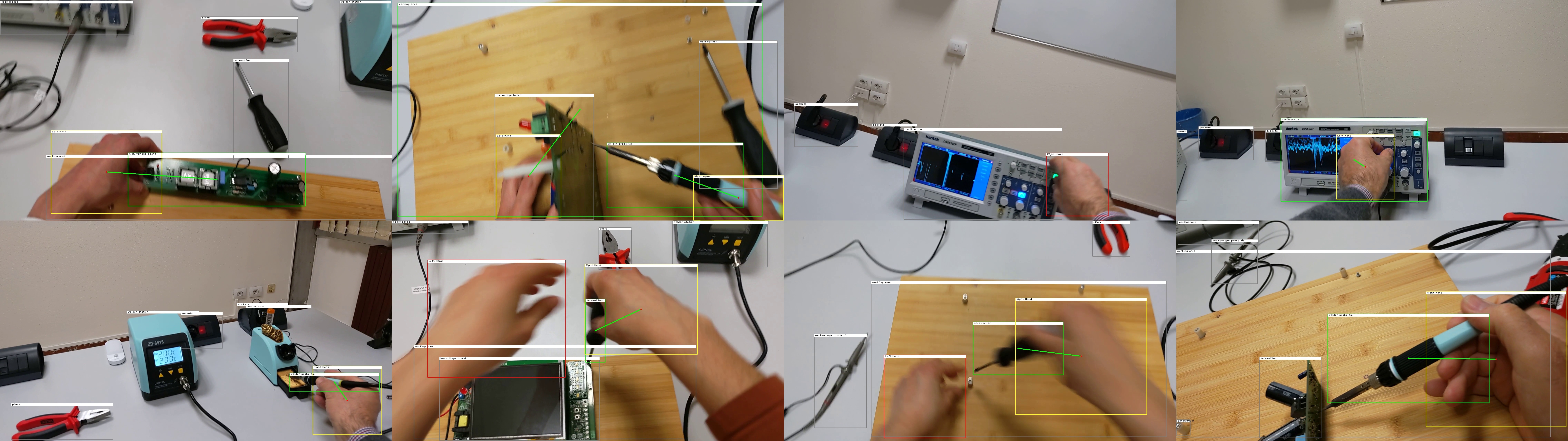}
    \caption{Qualitative results of the proposed method pretrained with the synthetic dataset and finetuned with the 100\% of the real data.} 
    \label{fig_qualitative_examples_proposed}
\end{figure}

\begin{figure}[t]
  \centering
  \begin{subfigure}[b]{0.40\linewidth}
    \includegraphics[width=\linewidth]{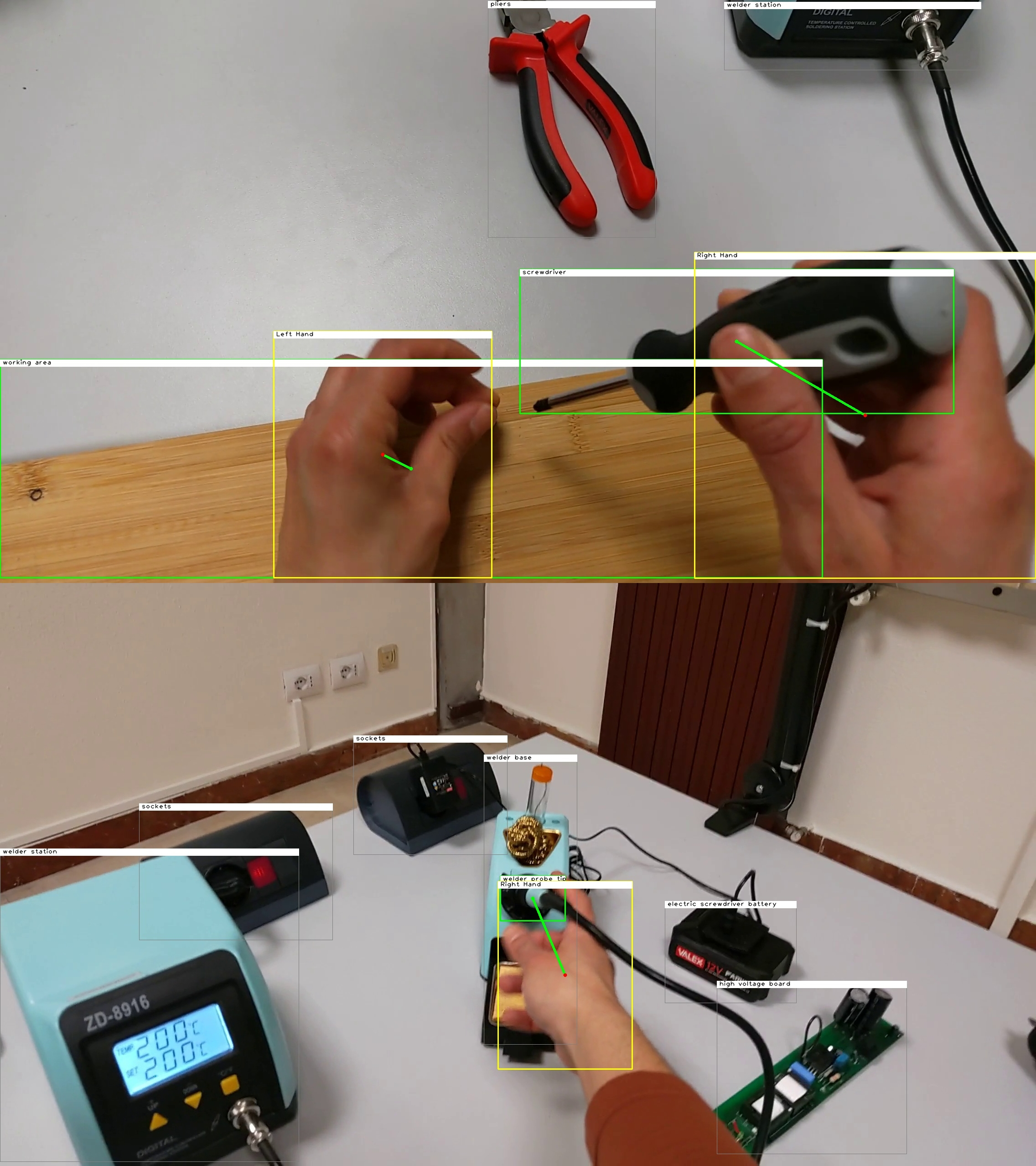}
    \caption{Proposed system}
  \end{subfigure}
  \begin{subfigure}[b]{0.40\linewidth}
    \includegraphics[width=\linewidth]{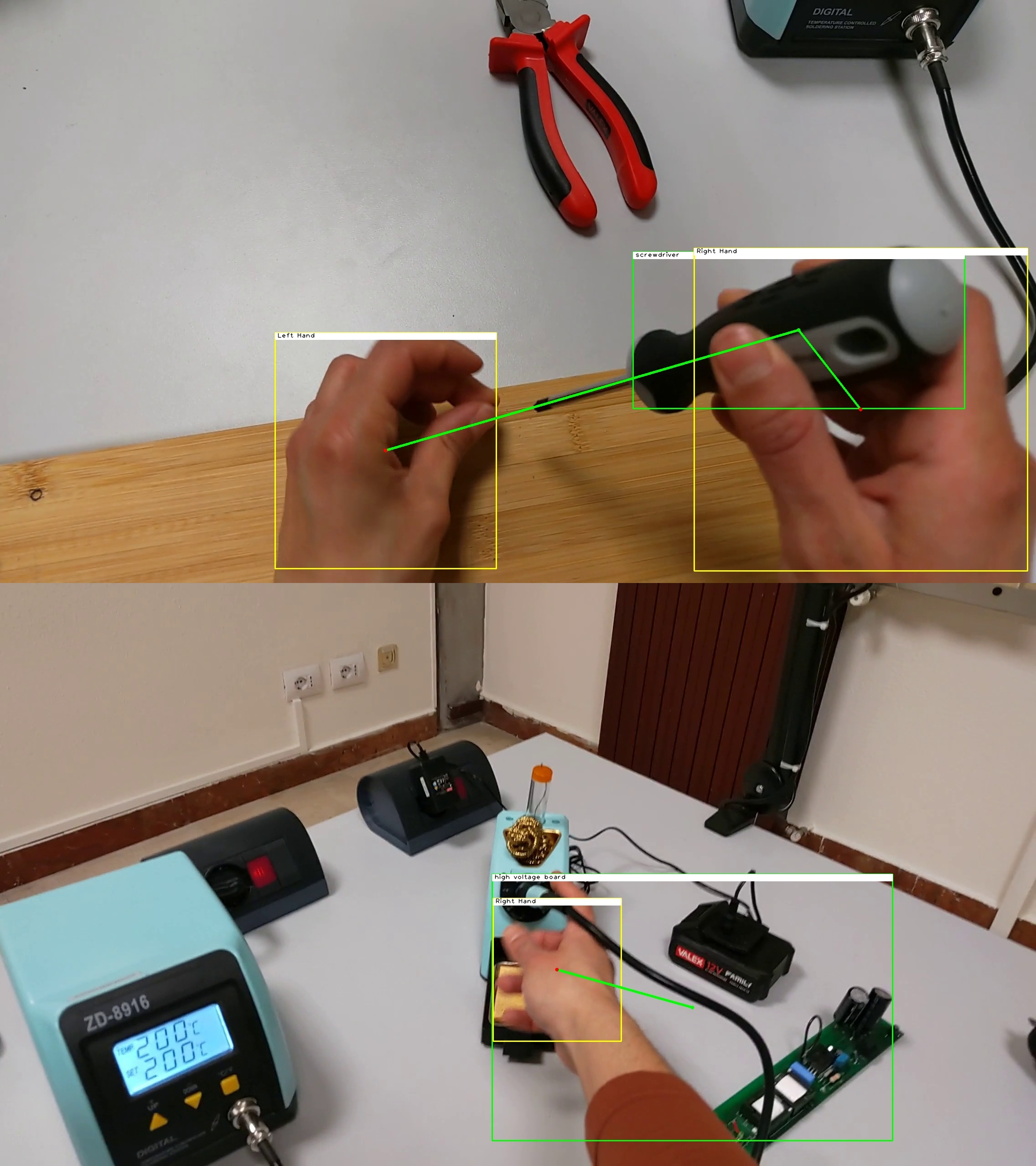}
    \caption{Method of~\cite{Shan2020UnderstandingHH} (BS5)}
  \end{subfigure}
  \caption{Comparison between our method trained with synthetic data and 100\% of the real dataset (a) and BS5 based on~\cite{Shan2020UnderstandingHH} (b).}
  \label{fig_comparative_examples_hic_proposed}
\end{figure}

Figure \ref{fig_graphs} - (b) shows the ``\textit{mAP All}'' curves related to the different models trained using synthetic data and different amount of real data (0\%, 10\%, 25\%, 50\%, 100\%). Figure~\ref{fig_qualitative_examples_proposed} reports some qualitative examples obtained using the model pretrained on the synthetic dataset and finetuned with $100\%$ of the real data. A qualitative comparison between the proposed method and~\cite{Shan2020UnderstandingHH} is shown in Figure~\ref{fig_comparative_examples_hic_proposed}.

\subsection{Additional experiment}
Table \ref{tab_comparison_proposed_system_only_active_object} shows an experiment in which the proposed method was trained using only the annotations of the active objects. From the results obtained is clear that including all objects in the object detection phase helps to obtain better overall results (\textit{mAP All} of $32.61\%$).
\begin{table}[t]
    \caption{Comparison between the models trained to recognize all the objects (\textit{EHOI\_S+R}) or only the active objects (\textit{EHOI\_ACTIVE\_S+R}).}
    \label{tab_comparison_proposed_system_only_active_object}
    \centering
    \resizebox{0.95\textwidth}{!}{%
    \begin{tabular}{l|c|c|c|c|c|c|c}
    \textbf{Model} & \textbf{Pretraining} & \textbf{Real Data\%} & \textbf{AP Hand} & \textbf{mAP Obj} & \textbf{mAR Obj} & \textbf{mAP H+Obj} & \textbf{mAP All} \\ \hline
    \textbf{EHOI\_S+R} & Syntehtic & 100 & \textbf{90.67} & 35.43 & \textbf{49.03} & \textbf{34.09} & \textbf{32.61} \\
    \textbf{EHOI\_ACTIVE\_S+R} & Syntehtic & 100 & \textbf{90.67} & \textbf{37.20} & 48.44 & 29.50 & 27.68 \\ \hline
    \end{tabular}%
}
\end{table}

\clearpage
\bibliographystyle{splncs04}
\bibliography{bibliography.bib}

\end{document}